\documentclass[lettersize,journal]{IEEEtran}
\usepackage{amsmath,amsfonts}
\usepackage{algorithmic}
\usepackage{algorithm}
\usepackage{array}
\usepackage[caption=false,font=normalsize,labelfont=sf,textfont=sf]{subfig}
\usepackage{textcomp}
\usepackage{stfloats}
\usepackage{url}
\usepackage{verbatim}
\usepackage{graphicx}
\usepackage{cite}
\usepackage{multirow}
\usepackage{amsmath,amsfonts}
\usepackage{mathrsfs}
\usepackage{stfloats}
\usepackage{diagbox}
\usepackage{bm}
\usepackage{booktabs}
\usepackage{soul, color, xcolor}
\soulregister\cite7
\soulregister\ref7
\soulregister\citep7 
\soulregister\citet7 
\soulregister\pageref7 

\graphicspath{ {./Figures} }

\begin{document}

\title{STBA: Towards Evaluating the Robustness of DNNs for Query-Limited Black-box Scenario}

\author{Renyang~Liu,~\IEEEmembership{Graduate Student Member,~IEEE,}
        Kwok-Yan Lam,~\IEEEmembership{Senior Member,~IEEE,}
        Wei~Zhou*,~\IEEEmembership{Member,~IEEE,}
        Sixing~Wu, 
        Jun Zhao,~\IEEEmembership{Member,~IEEE,}
        Dongting~Hu,
        Mingming~Gong*,~\IEEEmembership{Member,~IEEE,}

\IEEEcompsocitemizethanks{
\IEEEcompsocthanksitem{This work is supported in part by the National Natural Science Foundation of China under Grant 62162067 and 62101480, and in part by the National Research Foundation, Singapore and Infocomm Media Development Authority under its Trust Tech Funding Initiative, ABC Pte Ltd and XYZ association.}
\IEEEcompsocthanksitem R. Liu is with the School of Information Science and Engineering, Yunnan University, Kunming 650500, China (e-mail: ryliu@nus.edu.sg)
\IEEEcompsocthanksitem K.Y. Lam and J. Zhao are with the College of Computing and Data Science, Nanyang Technological University, Singapore, 639798 (e-mail: \{kwokyan.lam,junzhao\}@ntu.edu.sg).
\IEEEcompsocthanksitem W. Zhou and S. Wu are with the School of Software and the Engineering Research Center of Cyberspace, Yunnan University, Kunming 650500, China (e-mail: \{zwei, wusixing\}@ynu.edu.cn).
\IEEEcompsocthanksitem D. hu and M. Gong are with Melbourne Centre for Data Science, School of Mathematics and Statistics, University of Melbourne, Parkville, VIC 3010, Australia (e-mail: dongting@student.unimelb.edu.au; Mingming.gong@unimelb.edu.au).
\IEEEcompsocthanksitem *Corresponding authors: Wei Zhou, Mingming Gong.
}
}

\markboth{Journal of \LaTeX\ Class Files,~Vol.~14, No.~8, August~2021}%
{Shell \MakeLowercase{\textit{et al.}}: A Sample Article Using IEEEtran.cls for IEEE Journals}


\maketitle

\begin{abstract}
Extensive studies have revealed that deep neural networks (DNNs) are vulnerable to adversarial attacks, especially black-box {ones}, which can heavily threaten the DNNs deployed in the real world. Many attack techniques have been proposed to explore the vulnerability of DNNs and further help to improve their robustness. Despite the significant progress made recently, existing black-box attack methods still suffer from unsatisfactory performance due to the vast number of queries needed to optimize desired perturbations. Besides, the other critical challenge is that adversarial examples built in a noise-adding manner are abnormal and struggle to successfully attack robust models, whose robustness is enhanced by adversarial training against small perturbations. There is no doubt that these two issues mentioned above will significantly increase the risk of exposure and result in a failure to dig deeply into the vulnerability of DNNs. Hence, it is necessary to evaluate DNNs' fragility sufficiently under query-limited settings in a non-additional way. In this paper, we propose the Spatial Transform Black-box Attack (STBA), a novel framework to craft formidable adversarial examples in the query-limited scenario. {Specifically, STBA introduces a flow field to the high-frequency part of clean images to generate adversarial examples and adopts the following two processes to enhance their naturalness and significantly improve the query efficiency: a) we apply an estimated flow field to the high-frequency part of clean images to generate adversarial examples instead of introducing external noise to the benign image, and b) we leverage an efficient gradient estimation method based on a batch of samples to optimize such an ideal flow field under query-limited settings. Compared to existing score-based black-box baselines, extensive experiments indicated that STBA could effectively improve the imperceptibility of the adversarial examples and remarkably boost the attack success rate under query-limited settings.}
\end{abstract}


\begin{figure}[htp]
    \centering
    \setlength{\abovecaptionskip}{-5pt}
    \includegraphics[width=0.49\textwidth]
    {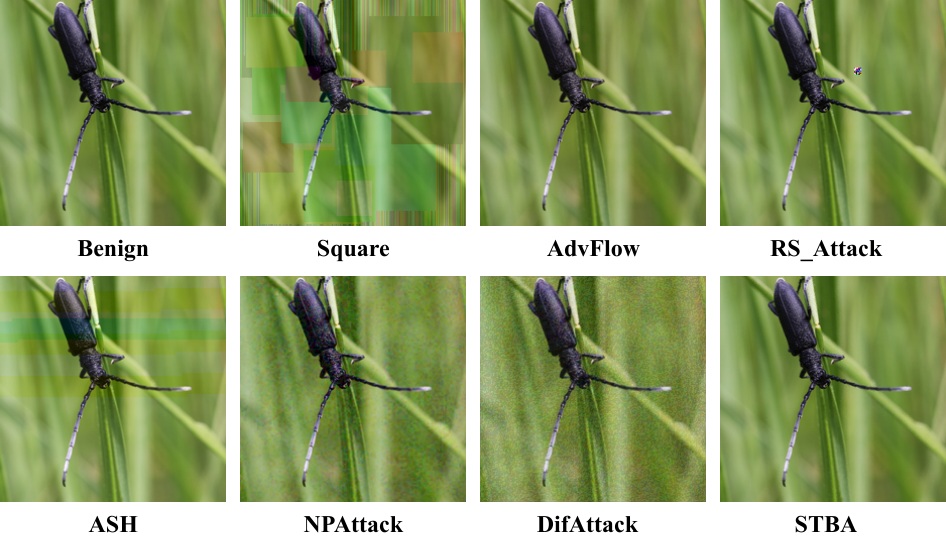}
    \caption{The clean example and their corresponding adversarial example generated by baselines and the proposed STBA. The first one is the clean images, and the flowings are the adversarial example generated by Square Attack \cite{eccv/AndriushchenkoC20}, AdvFlow \cite{nips/DolatabadiEL20}, RS\_Attack \cite{aaai/CroceASF022}, ASH \cite{mir/LiH00S22}, NPAttack \cite{pr/BaiWZJX23}, and DifAttack \cite{aaai/00710ZT24}, respectively.}
    \label{fig:visual}
    \vspace{-0.5cm}
    
\end{figure}

\section{Introduction}
\label{sec:intro}

\IEEEPARstart{D}{eep} neural networks (DNNs) are susceptible to adversarial examples \cite{tkdd/XiongWZCLZH22,corr/abs-2309-13857,tdsc/HeWLYJLZ23}, which are crafted by subtly perturbing a clean input \cite{tmm/XingYLLZW23,ijautcomp/XuMLDLTJ20,tmm/ZhangWLSZ22,tmm/YuanCZZLY23}, especially for computer vision (CV) \cite{tmm/abs-2212-01628} and Natural Language Processing (NLP) tasks, such as image classification, face recognition \cite{tist/TuMZDXF20}, and language translation \cite{tmm/TangGH022}. Except for the threat brought by such attacks, adversarial examples are the most effective way to evaluate existing models' vulnerability and can further impel the defender to improve the robustness of existing models and guide the design of new safety models. The critical point in carrying out adversarial attacks on CV models is how to generate adversarial examples with high imperceptibility and attack success rate. Various methods have been proposed to build adversarial examples; among them, the black-box attacks are more practical because they do not need to access the target model and only require the prediction results \cite{tifs/LiuZZCZL24,cvpr/WangYY0FDLHX21,tmm/DuP22,eaai/ZhangZLZSZ22}, which makes them are more suitable to destroy the DNN models deployed in the physical world.

Although most existing black-box attacks can obtain a high success rate by adding noise to the original image, they are not ideal in terms of naturalness since the added perturbations are not harmonious with the original clean image \cite{iclr/MadryMSTV18}. Meanwhile, most existing black-box attacks are query and optimization-based, which regard the process of calculating adversarial examples (or adversarial noise) as an optimization problem. They iteratively optimize the intermediate {adversarial example with} the query results until it can attack the target model successfully \cite{ccs/ChenZSYH17}. Those methods require a vast number of queries to ensure that most crafted adversarial examples for a dataset can be used to attack victim models successfully, which results in a heavy computation and time burden; worse, a mass of queries can be easily detected by defenders. Finally, most deployed DNN models are enhanced by various defense mechanisms; among them, adversarial training, which finetunes the DNN model on the pre-collected adversarial examples, is verified as the most effective way to improve the model's robustness against unknown attacks. Since such adversarial examples are often locally generated with noise-adding attacks, existing black-box attacks are more challenging to {attack these robust models successfully.}

To address the issues mentioned above, researchers have proposed various works to solve them separately. For improving the imperceptibility of adversarial examples \cite{tmm/ZhangZYZ23,tmm/ChengGJLFLL22}, some methods try to generate imperceptible adversarial examples, such as AdvFlow \cite{nips/DolatabadiEL20}, which crafts adversarial examples by disturbing the latent space of image under $L_p$-norm constrained. Even though this method makes the adversarial perturbations more imperceptible to human eyes under black-box settings, it still impacts the whole clean image. For the challenge of a large number of queries required by optimized-based methods, some query-limited or query-efficient black-box attack methods, like evolutionary-based methods $\mathcal{N}$attack \cite{icml/LiLWZG19} and gradient estimation-based AdvFlow \cite{nips/DolatabadiEL20}, declare that they can build suitable adversarial examples only in thousands or hundreds of queries. However, existing methods still need a large query budget to pursue a higher attack success rate. The empirical maximum query budget is usually 10,000 $ \sim $ 100,000 \cite{iclr/BrendelRB18, icml/GuoGYWW19}, which is still intolerable to online deployed DNN systems. As a consequence, a large number of queries yield an increased query cost and the probability of exposure. 

To better evaluate the vulnerability and robustness of existing DNN models, including the robust ones, and improve the concealment of adversarial examples and the query efficiency of black-box attacks under the query-limited scenario, we formulate the issue of synthesizing adversarial examples beyond additive perturbations and propose a novel non-addition black-box attack method called \textbf{STBA}. More specifically, STBA uses spatial transformation techniques to generate adversarial examples with better concealment rather than directly adding well-designed noise to the benign image. The spatial transform technique can generate an eligible adversarial example by leveraging a well-optimized smooth flow field matrix $\bm{f}$ to move each pixel to a new location. {Besides, to further enhance the imperceptibility of the generated adversarial noise, we first split the high- and low-frequency parts from clean images and then only apply the will-estimated flow field $\bm{f}$ to the high-frequency part to formulate the adversarial high-frequency counterpart, which will be combined to the original low-frequency part to synthesis the final adversarial examples.} Furthermore, to address the query challenge of optimization-based black-box attacks, STBA leverages the scarce information, query accounts, attack success rate, and the risk of exposure in the query-limited situation, employed with gradient estimation to optimize such a flow field, which can effectively find the global optima with higher probability but require less information from the target model and straightforward the generation process. Therefore, the proposed STBA can achieve a higher attack success rate with a harsher query budget.

We draw the adversarial examples of different methods in Fig. \ref{fig:visual}, where the victim model is ResNet-152 \cite{cvpr/HeZRS16} and the image is from the ImageNet \cite{cvpr/DengDSLL009} dataset. The visual results show that our proposed method alters the clean image slightly, thus ensuring the invisibility of the adversarial perturbations. Furthermore, we conducted extensive experiments on four different CV benchmark datasets, the statistical results indicate that the STBA can obtain the highest attack success rate with the max query budget limited as $Q_{max}=1000$ or $Q_{max}=10000$. Specifically, STBA can achieve a nearly 100\% attack success rate on the normally trained models when $Q_{max}=1000$ and on the robust trained models when $Q_{max}=10000$ in most cases, while the baselines are lower than 90\% and {70\%} in the most cases, respectively. Furthermore, we found that STBA can achieve the highest attack success rate with fewer average queries compared to the existing query-limited black-box methods on the target victim models while outperforming them concerning attack success rate. The main contributions could be summarized as follows:

\begin{itemize} 
    \item We propose a new black-box attack approach called STBA, which introduces the spatial transform techniques instead of noising-adding to black-box settings to address the low attack success rate and query efficiency of existing black-box methods on normally and robustly trained DNNs in query-limited scenarios.  

    \item To further enhance the imperceptibility of adversarial examples, we integrate Gaussian blur techniques and dynamical strategy into our attack framework. Specifically, we first decomposed the image into high-frequency and low-frequency components. We then only applied the dynamically constrained flow field $\bm{f}$ to the high-frequency part to improve the generated adversarial images' visibility and image quality.
  
    \item Comparing with the existing score-based black-box methods, experimental results on four benchmark datasets show STBA's superiority in synthesizing adversarial examples with efficient queries, attack performance, better naturalness, and remarkable transferability.
\end{itemize}

The rest of this paper is organized as follows. We briefly review the methods relating to adversarial attacks in Sec. \ref{Sec:related}. In Sec. \ref{sec:methodology}, we provide the details of the proposed STBA framework. The experiments are presented in Sec. \ref{Sec:experiments}, with the conclusion and limitation drawn in Sec. \ref{Sec:conclusion}.

\section{Related Work}
\label{Sec:related}
This section briefly reviews the most relevant attack methods to the proposed work, including spatial transform-based attacks and Query-Efficient black-box attacks.

\subsection{Spatial transform-based attacks}
The concept of spatial transformation is first mentioned in \cite{nips/JaderbergSZK15}, which indicates that conventional neural networks are not robust to rotation, translation, and dilation. Next, \cite{iclr/XiaoZ0HLS18} utilized the spatial transform techniques and proposed the stAdv method to craft adversarial examples with a high fooling rate and perceptually realistic beyond noise-adding way. They change each pixel position in the clean image by applying a well-designed flow field matrix $\bm{f}$ to {the clean image to formulate the adversarial counterpart}. Later, \cite{icdm/ZhangRW020} proposed a new method to produce the universal adversarial examples by combining the spatial transform and {pixel-level} distortion, and it successfully increased the attack success rate against universal perturbation to more than 90\%. In the literature \cite{mm/AydinSKHT21}, the authors applied spatial transform to the YUV space to generate adversarial examples with higher superiority in image quality. Also, Liu \textit{et al}. \cite{finr/LiuJHZWZZ23} utilized spatial transformations by applying the flow field to the latent representations from the normalized flow model \cite{nips/KingmaD18} to guarantee that generated adversarial examples are more perceptually similar to the original image.

\subsection{Query-Efficient black-box attack}
Although most optimized-based black-box attacks have achieved a high attack success rate, even reaching 100\% in some cases, the tremendous need for queries to optimize adversarial examples is still a big challenge. Empirically, the average number of queries required to obtain a suitable adversarial example is from thousands to hundreds of thousands \cite{icml/GuoGYWW19, icml/IlyasEAL18}, which is obviously not feasible in the physical world and can be detected by the target DNN models easily. Therefore, recently, researchers have proposed various methods to generate adversarial examples under query-limited queries or in a query-efficient way \cite{iclr/ChengSCC0H20,cvpr/MaCY21}.

The first work to limit the number of queries in black-box attacks is the Bayesian Optimization-Based Black-box Attack \cite{corr/abs-1712-08713}. It uses the Bayesian optimization strategy to reduce the average number of queries to 1/10 of the random perturbation strategies. Then, Ilyas \textit{et al}. proposed NES \cite{icml/IlyasEAL18} to improve the query efficiency while maintaining a higher success rate in query and information-limited settings. In \cite{cvpr/LiXZYL20}, the authors propose a query-efficient boundary-based attack (QEBA) that only relies on the predicted hard labels and presents an optimal analysis of gradient estimation based on dimensional reduction and gets good results. AdvFlow \cite{nips/DolatabadiEL20} improves the attack performance with a limited query budget by disturbing the image's latent representation obtained from pre-trained Normalize Flow. Square Attack \cite{eccv/AndriushchenkoC20} uses the random searched square-like adversarial perturbation to extend attacks. Accelerated Sign Hunter (ASH) \cite{mir/LiH00S22} crafts the adversarial example by applying a Branch Prune Strategy that infers the unknown sign bits to avoid unnecessary queries and further improve the query efficiency within a limited query budget.

Besides, some researchers have proposed combining the transferability of adversarial attacks to improve query efficiency, which is dubbed query- and transfer-based attacks. F. Suya \textit{et al}. \cite{uss/SuyaC0020} combined transfer-based attack \cite{tmm/GaoHSYS22} and optimize-based black-box and proposed Hybrid Batch Attacks (HBA). The HBA can use the information of the existing local models to generate candidate adversarial seeds and use these seeds to query the target model and optimize the generated images iteratively. Its seed-first strategy saves about 70\% query times for an attack. MCG-Attack \cite{pami/YinZWFZFY24} improves the query efficiency by fully utilizing both example-level and model-level transferability. DifAttack \cite{aaai/00710ZT24} disentangles image latent into adversarial and visual features, trains an autoencoder with clean images and their adversarial examples of surrogate model(s), and iteratively optimizes adversarial features based on victim model feedback to create successful AEs while maintaining visual consistency.

\begin{figure*}[htp]
      \centering
      \includegraphics[width=0.93\textwidth]{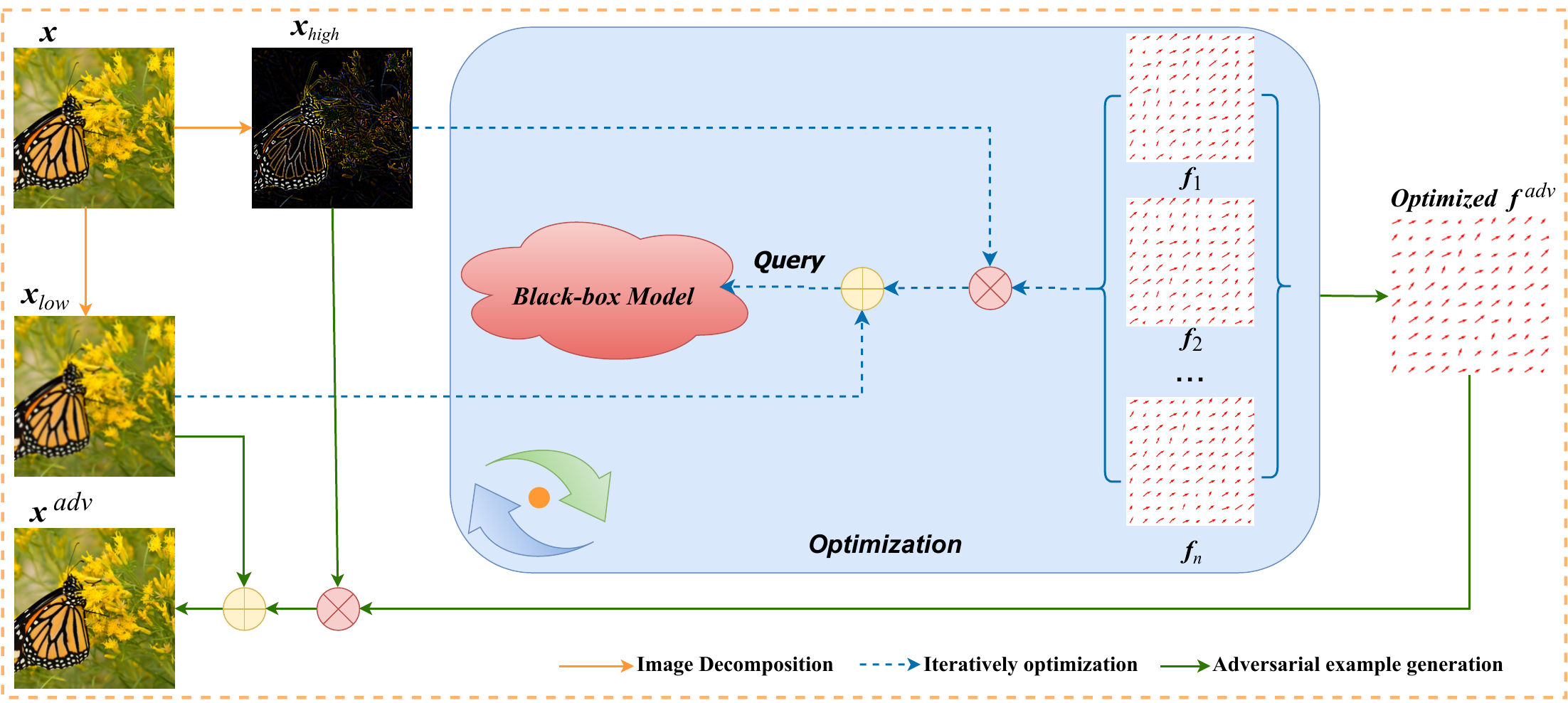}
      \caption{Overview of Spatial Black-box Transform Attack (STBA). The \textbf{Black-box Model} is the victim model. The $\bm{x}$ and $\bm{x}^{adv}$ are the benign image and the corresponding adversarial counterpart, respectively. The ${\bm{x}_{high}}$ and $\bm{x}_{low}$ represent the high-frequency part and the low-frequency part of the benign image. The $\otimes$ represents the spatial transformation operation, and $\oplus $ means element addition. \textbf{The $\bm{f}_i, i\in \{1,2,...,n\}$ are the mini-batch candidate flow field sampled from an isometric normal distribution $\mathcal{N}$}, and the $\bm{f}^{adv}$ is the final optimized flow field that applies to the original clean image to formulate the final adversarial image $\bm{x}^{adv}$.}
      \label{fig:framwork}
      \vspace{-0.4cm}
\end{figure*}
{However, these transfer-based attacks heavily rely on the similarity between the substitute model and the oracle model. In this paper, we only focus on score-based attacks, which are independent of local surrogate models.} Therefore, adversarial attacks, especially for the Query-Efficient black-box attacks, pose the request for a method that is efficient and effective {in building} adversarial examples with high imperceptibility and quality in performing attacks on unknown DNN models within limited queries. On the other hand, with the development of defense mechanisms, higher requirements are placed on the invisibility of adversarial examples. To achieve these goals, we learn from the previous studies that adversarial examples can be gained in a beyond-noise-adding way. {Hence, we are well motivated to develop a novel method to disturb the original image by applying a well-calculated flow field $\bm{f}$ to clean images to generate adversarial ones, which makes it possible to build qualified adversarial examples with high invisibility in dozens or hundreds of queries.}

\section{Methodology}
\label{sec:methodology}

\subsection{Problem Definition}
Given a well-trained DNN classifier $ \bm{\mathcal{C}} $ and an input $ \bm{x} $ with its corresponding label $ y $, we have $ \bm{\mathcal{C}}(\bm{x})=y $. The adversarial example $ \bm{x}^{adv} $ is a neighbor of $ \bm{x} $ and satisfies that $ \bm{\mathcal{C}}(\bm{x}^{adv}) \neq y $ and $ \left \| \bm{x}^{adv}-\bm{x} \right \|_p \leq \epsilon  $, where the $ \bm{L}_p $-norm is used as the distance measure and $ \epsilon $ is usually a small noise budget. With this definition, the problem of finding an adversarial example becomes a constrained optimization problem:
\begin{equation}
      \label{eq:eq1}
      \bm{x}^{adv}= 
\underset{\left \| \bm{x}^{adv}-\bm{x} \right \|_p \leq \epsilon}{\mathop{arg max}\mathcal{L}} ( \bm{\mathcal{C}}(\bm{x}^{adv}) \neq y),
\end{equation}
where $ \mathcal{L} $ stands for a loss function that measures the confidence of the model outputs.

Previous works craft an adversarial example $\bm{x}^{adv}$ by adding $ \bm{L}_p $-norm constrained noise to the clean image $\bm{x}$, i.e.,
\begin{equation}
    \label{eq:eq2}
    \bm{x}^{adv} = \bm{x}+noise, \  s.t. \  \left \| noise \right \|_p \leq \epsilon,
\end{equation}
Different from this, we introduce the spatial transform techniques {to the high-frequency part of the clean image to build the adversarial example $\bm{x}^{adv}$.} As illustrated in Fig. \ref{fig:framwork}, the proposed Spatial Transform Black-box Attack framework can be divided into flowing stages: {The first stage is decomposing the different frequency information, i.e., the high-frequency part ${\bm{x}_{high}}$ and the low-frequency part $x_{low}$, from the original clean image $\bm{x}$ with the help of Gaussian Blur (we set the kernel size as 3x3). Second, we random sampling a mini-batch candidate flow fields $\bm{f}_i, i\in \{1,2,3...,n\}$ from an isometric normal distribution $\mathcal{N}$, which will be applied to the high-frequency part ${\bm{x}_{high}}$, and then combined with the low-frequency part $\bm{x}_{low}$ to build the candidate adversarial examples used to query the black-box victim model to estimate the gradient in an iterative manner. Once the gradient is estimated, it can used to guide the optimization process to update the flow field $\bm{f}^{adv}$. Finally, we can obtain the adversarial example $\bm{x}^{adv}$ by applying the calculated flow field $ \bm{f}^{adv} $ to the high-frequency part ${\bm{x}_{high}}$.}

\subsection{Adversarial Example Generation}
\label{subsec:adversarial}
We utilize the spatial transform method to build adversarial examples in a non-noise additional way. The spatial transform techniques using a flow field matrix $\bm{f} \in [0,1]^{2*h*w}$ to transform the original image $\bm{x}$ to $ \bm{x}_{st}$ \cite{iclr/XiaoZ0HLS18}, where $h$ and $w$ are the height and width of the image. In particular, for our proposed STBA, we only need to apply such transform operation to the high-frequency part, i.e., ${\bm{x}_{high}}$. Specifically, assume the input is ${\bm{x}_{high}}$ and its transformed counterpart $ \bm{x}_{high}^{st}$, for the $i$-th pixel in $\bm{x}_{high}^{st}$ at the pixel location $ (u_{st}^i,v_{st}^i) $, we need to calculate each element of the flow field as $\bm{f}_i = (\Delta u^i, \Delta v^i) $. So, the $i$-th pixel $ \bm{x}_{high}^i$'s location in the transformed image can be indicated as:
\begin{equation}
      \label{eq:eq3}
      (u^i,v^i) =(u_{st}^i + \Delta u^i ,v_{st}^i+ \Delta v^i ).
\end{equation}

To ensure the flow field $ \bm{f} $ is differentiable, the bi-linear interpolation \cite{nips/JaderbergSZK15} is used to obtain the 4 neighboring pixels' value surrounding the location $(u_{st}^i + \Delta u^i, v_{st}^i+ \Delta v^i )$ for the transformed image $\bm{x}_{high}^{st}$ as:
\begin{equation}
      \label{eq:eq4}
      {\bm{x}_{high}^{st}}^i=\sum_{q \in \bm{N} (u^i,v^i)}\bm{x}_{high}^q(1-|u^i-u^q|) (1-|v^i-v^q|),
\end{equation}
where $ \bm{N} (u^i,v^i) $ is the neighborhood, that is, the four positions (top-left, top-right, bottom-left, bottom-right) tightly surrounding the target pixel $ (u^i,v^i) $. For each pixel ${\bm{x}_{high}^{st}}^i$ in the adversarial high-frequency part $\bm{x}_{high}$, we can calculate it with with Eq. \ref{eq:eq4} to formulate the $\bm{x}_{high}^{st}$, i.e., the final adversarial high-frequency part $\bm{x}_{high}^{adv}$. Once the $\bm{f}$ for formulating $\bm{x}_{high}^{adv}$ has been computed, we can obtain the final $\bm{x}^{adv}$ by adding such adversarial high-frequency part $\bm{x}_{high}^{adv}$ with low-frequency part $\bm{x}_{low}$ of the clean image, which is given by:

\begin{equation}
    \label{eq:eq5}
    \bm{x}^{adv}=\bm{x}_{high}^{adv} \oplus \bm{x}_{low}.
\end{equation}


In practice, we regard the problem of calculating flow field $\bm{f}$ as an optimization task. Under the black-box settings, the only information that we can utilize is the output of the classifier $ \bm{\mathcal{C}} (\cdot) $ regarding a specific input. To this end, following the previous works \cite{icml/LiLWZG19,nips/DolatabadiEL20}, in this paper, we use gradient estimation to find the optimal flow $\bm{f}$ in an iterative manner.

\subsection{Objective Functions}
Taking the attack success rate and visual invisibility of the generated adversarial examples into account, we divide the objective function into two parts, where one is the adversarial loss $\mathcal{L}_{adv}$ and the other is a constraint for the flow field, i.e., flow loss $\mathcal{L}_{\bm{f}}$. 

For adversarial attacks, the goal is making $\bm{\mathcal{C}}(\bm{x}^{adv}) \neq y$. We give the objective function as follows:
\begin{equation}
    \label{eq:eq6}
     \mathcal{L}_{adv} (\bm{x}^{adv},y,\bm{f})= max[\bm{\mathcal{C}}(\bm{x}^{adv})_{y}-\underset{k\neq y}{max}\bm{\mathcal{C}}(\bm{x}^{adv})_{k},k].
\end{equation}

To guarantee the visual imperceptibility of the generated adversarial examples, we extend the flow loss defined in \cite{iclr/XiaoZ0HLS18} as $\mathcal{L}_{f}$ to constrain the distance of any two adjacent pixels, which compute the sum of spatial movement as:
\begin{multline}
    \mathcal{L}_{\bm{f}} (\bm{f})= \\    
    \sum_{p}^{A} \sum_{q \in \mathcal{N}(p) }^{} 
\sqrt{\left \| \bigtriangleup u^{(p)} - \bigtriangleup u^{(q)} \right \| _{2}^{2}
 + \left \| \bigtriangleup v^{(p)} - \bigtriangleup v^{(q)} \right \| _{2}^{2} }, 
\end{multline}
where $A$ represents all the pixels in the victim image.

In addition, we introduce a dynamic adjustment strategy for updating the flow field $\bm{f}$ flexible to further improve the generated adversarial examples' invisibility. More specifically, we initialize the flow budget $\xi_{t_0}$ with a small value (such as $\xi_{t_0}=10^{-1}$), then we update $\xi$ with $T$ steps, that is:

\begin{multline}
    \label{eq:xi_update}
     \xi_{t}=\xi_{t-1}+\alpha, \ \alpha = (\xi_{max}-\xi_{t_0}) / T, \\
     s.t. \  t\in\{1,2,3,...,T\},  
\end{multline}
where $T=Q_{max} / n_{sample} $, and the value of $\bm{f}$ matrices of each iterations can be obtained with a $clip(\cdot )$ operation, i.e., $\bm{f}=clip(\bm{f},-\xi ,+\xi )$.

The whole objective function can be written as:
\begin{equation}
    \label{eq:loss_total}
    \mathcal{L} = \mathcal{L}_{adv} + \lambda* \mathcal{L}_{\bm{f}}, \ s.t. \ \bm{f}=clip(\bm{f},-\xi,+\xi),  
\end{equation}
where $\lambda$ is a hyper-parameter to balance each loss item, in this paper, we set it as $\lambda=5$ {(For the reason why we set $\lambda=5$, please refer to Sec. \ref{sec:ablation} ).}

\subsection{Optimization}
Recall that our goal is to calculate the flow field $\bm{f}$, which will be applied to the clean image's high-frequency part $\bm{x}_{high}$ to generate its adversarial counterpart one $\bm{x}_{high}^{adv}$ and further formulate adversarial example $\bm{x}^{adv}$. we regard the flow field $\bm{f}$ comes from a unique distribution and re-defined the expected adversarial example $\bm{x}^{adv}$ by re-written the Eq.\ref{eq:eq4} and Eq.\ref{eq:eq5} as 
\begin{equation}
    \bm{x}^{adv}=\bm{x}_{high} \otimes \bm{f} \oplus \bm{x}_{low} \  s.t. \  \bm{f} \sim \mathcal{N}(\bm{f}|\bm{\mu},\sigma^2\bm{I}),
\end{equation}
where $\mathcal{N}$ is an isometric normal distribution with mean $\bm{\mu}$ and standard deviation $\sigma$, $\otimes$ stands for the spatial transform operation. As in AdvFlow \cite{nips/DolatabadiEL20}, we flowing the \textit{lazy statistician} \cite{wasserman2004all} and re-write our attack definition in Eq. \ref{eq:loss_total} as
\begin{align}
    J(\bm{\mu},\sigma)&= \mathbb{E}_{p(\bm{x}^{adv}|\bm{\mu},\sigma)}[\mathcal{L}(\bm{x}^{adv})] \nonumber \\
    &= \mathbb{E}_{\mathcal{N}(\bm{f}|\bm{\mu},\sigma^{2} \bm{I})}[\mathcal{L}(\bm{x}_{high} \otimes \bm{f} \oplus \bm{x}_{low})],
\end{align}
Further, we only required to minimize $J(\bm{\mu},\sigma)$ w.r.t $\bm{\mu}$ \cite{icml/LiLWZG19,nips/DolatabadiEL20}. Likely, we derive the Jacobian of $J(\bm{\mu},\sigma)$ as 
\begin{align}
    \nabla J(\bm{\mu},\sigma )&=\mathbb{E}_{p(\bm{x}^{adv}|\bm{\mu},\sigma^{2}\bm{I})} \nonumber \\ 
    &[\mathcal{L}(\bm{x}_{high} \otimes \bm{f} \oplus \bm{x}_{low}) \nabla _{\bm{\mu}}log \mathcal{N}(\bm{f}|\bm{\mu},\sigma^2\bm{I}) ],
\end{align}
The value of expectation can be obtained by sampling from a distribution $\mathcal{N}(\bm{f}|\bm{\mu},\sigma^2\bm{I})$ and the $\bm{\mu}$ can be optimized by executing a gradient descent step
\begin{equation}
    \bm{\mu} \longleftarrow \bm{\mu} - lr \cdot \nabla_{\bm{\mu}} \bm{J}(\bm{\mu},\sigma ), 
\end{equation}
where the $lr$ is the learning rate. We find the optimal flow field $\bm{f}$, which will lead the clean example $\bm{x}$ change to adversarial, by iteratively searching the optimization direction on a mini-batch-wised random sampled neighbor of $\bm{f}$ from $\mathcal{N}(\bm{f}|\bm{\mu},\sigma^2\bm{I})$, here, we define the batch-size as $B=n_{sample}$. Finally, the adversarial example $\bm{x}^{adv}$ can be derived as 
\begin{equation}
\label{eq:x_adv_final}
    \bm{x}^{adv}=\bm{x}_{high} \otimes clip(\bm{f}^{adv},-\xi,\xi) \oplus \bm{x}_{low}.
\end{equation}

The whole algorithm of STBA is listed in Alg. \ref{alg:alg1} for ease of reproduction of our results, where lines 4-20 depict the core optimization process.

\begin{algorithm}[htp]
    \caption{Spatial Transform for Query-Efficient Attack}
    \label{alg:alg1}
    \textbf{Input}: $ \bm{\mathcal{C}} $: the target classifier to be attacked;
		$ \bm{x} $: the clean image;
		$ \bm{Q}_{max} $: the maximum querying number;
		$ q $: the current querying number;
        $ \xi $: the flow budget;
        $lr$: the learning rate;
        $n_{sample}$: the number of samples;
        $adjust_{num}$: the epoch to update the flow budget.\\
    \textbf{Parameter}: The flow field $ \bm{f} = [0, 1]^{2*h*w} $.\\
    \textbf{Output}: The adversarial example $\bm{x}^{adv}$ used for attack.
    \begin{algorithmic}[1] 
    \STATE Initialize the flow budget $ \xi $ with $ \xi=10^{-1}$.
    \STATE Initialize $ \bm{\mu} $ randomly.
    \STATE Initialize the parameters of the flow $ \bm{f_0} $ with zeros.
    \STATE Decompose $\bm{x}$ to $\bm{x}_{high}$ and $\bm{x}_{low}$.
    \WHILE{$q<=\bm{Q}_{max}/n_{sample}$}      
        \STATE Draw $n_{sample}$ samples from $\bm{\delta}_{\bm{f}} = \bm{\mu} + \sigma \cdot \bm{\epsilon}$ s.t. $\bm{\epsilon} \sim \bm{\mathcal{N}}(\bm{\epsilon} | \bm{0}, \bm{I})$.
        \STATE Set $\bm{f}_k = \bm{\delta_{f_k}}$ for all $k = 1,...,n_{sample}$.
        \STATE Compute $ {\bm{x}_{high}^{adv}}^k $ by Eq. \ref{eq:x_adv_final} for all $k = 1,...,n_{sample}$.
        \STATE Calculate $\bm{\mathcal{L}_k} = \bm{\mathcal{L}}({\bm{x}_{high}^{adv}}^k \oplus \bm{x}_{low})$ by Eq. \ref{eq:loss_total} for all $k = 1,...,n_{sample}$.        
        \STATE Normalize $\hat{\bm{\mathcal{L}}}_k = (\bm{\mathcal{L}}_k - mean(\bm{\mathcal{L}}))/std(\bm{\mathcal{L}})$.
        
        \STATE Compute $\nabla_{\bm{\mu}} \bm{J}(\bm{\mu},\sigma ) = \frac{1}{n_{sample}}  {\textstyle \sum_{k=1}^{n_{sample}}}\hat{\mathcal{L} } \cdot \bm{\epsilon }_k $.
        
        \STATE update $\bm{\mu} \longleftarrow \bm{\mu} - lr \nabla_{\bm{\mu}} \bm{J}(\bm{\mu},\sigma ) $.        
       
        \IF{$q \  \% \  adjust_{num} = 0$}
            \STATE Update $\xi_{t}$ by Eq.\ref{eq:xi_update}.
        \ENDIF
        \STATE Compute $\bm{x}^{adv}_{t}=\bm{x} \otimes clip(\bm{f}_0+\bm{\mu},-\xi_t,\xi_t) \oplus \bm{x}_{low} $.
        \IF{$ \bm{x}^{adv}_{t} $ attack $\bm{\mathcal{C}}$ successfully}
            \STATE break.
        \ENDIF
        
    \ENDWHILE
    \STATE \textbf{return} $ \bm{x}^{adv}_{t} $
    \end{algorithmic}
\end{algorithm}

\begin{table*}[]
\caption{\textbf{The victim models and their corresponding classification accuracy (\%)} on the validation set of CIFAR-10, CIFAR-100, STL-10 and the subset of ImageNet dataset NIPS2017, respectively.}
\label{tab:victim}
\small
\centering
\renewcommand{\arraystretch}{1.1}
\setlength\tabcolsep{1.5pt}

\begin{tabular}{c|lc|lc|lc|lc}
\toprule
\multirow{2}{*}{Type}   & \multicolumn{2}{c|}{CIFAR-10}     & \multicolumn{2}{c|}{CIFAR-100}    & \multicolumn{2}{c|}{STL-10} & \multicolumn{2}{c}{ImageNet}      \\
                        & Model Name                   & Accuracy & Model Name                   & Accuracy & Model Name           & Accuracy   & Model Name                   & Accuracy \\ \midrule
\multirow{4}{*}{Normal} & VGG-19                 & 93.91    & VGG-19                 & 73.84    & VGG-19         & 75.67      & VGG-19                 & 89.00    \\
                        & ResNet-56              & 94.38    & ResNet-56              & 72.60    & ResNet-56      & 84.58      & ResNet-152              & 94.00    \\
                        & MobileNetV2            & 93.12    & MobileNetV2            & 71.13    & MobileNetV2    & 78.56      & MobileNetV2            & 87.80    \\
                        & ShuffleNetV2           & 93.98    & ShuffleNetV2           & 75.49    & ShuffleNetV2   & 79.46      & DenseNet-121           & 70.20    \\ \midrule
\multirow{4}{*}{Robust} & Sehwag2020Hydra        & 88.98    & Pang2022Robustness     & 65.56    & FreeAT           & 63.77      & Salman2020Do\_R50      & 77.80    \\
                        & Wang2020Improving      & 87.50    & Addepalli2022Efficient & 65.45    & FastAdv           & 65.65      & Engstrom2019Robustness & 77.40    \\
                        & Zhang2019Theoretically & 84.92    & Sehwag2021Proxy        & 65.93    & TRADES         & 74.10      & Wong2020Fast           & 62.70    \\
                        & Wong2020Fast           & 83.33    & Cui2020Learnable       & 70.24    & MART           & 71.14      & Salman2020Do\_R18      & 63.10    \\ \bottomrule
\end{tabular}
\end{table*}
\vspace{-10pt}

\begin{table*}[!htp]
\caption{\textbf{Attack success rate (ASR,\%) against normal models} on four benchmark datasets under query budget is set to \textbf{1000} and \textbf{10000}}
\label{tab:clean}
\small
\centering
\renewcommand{\arraystretch}{1.1}
\setlength\tabcolsep{2.5pt}
\begin{tabular}{cccccccccccccccc}
\toprule
\multicolumn{2}{c}{Max$_Q$}               & \multicolumn{7}{c}{1000}                                                       & \multicolumn{7}{c}{10000}     \\
\cmidrule(lr){1-2} \cmidrule(lr){3-9}  \cmidrule(lr){10-16}
Dataset                    & Model        & Square & AdvFlow & RS & ASH   & NP & Dif & STBA            & Square          & AdvFlow & RS & ASH   & NP        & Dif       & STBA            \\ \midrule
\multirow{4}{*}{CIFAR-10}  & VGG-19       & 90.27  & 64.71   & 88.99      & 96.47 & 90.91    & 96.89     & \textbf{99.64}  & 99.65           & 97.05   & 91.38      & 96.51 & 99.70           & 99.97           & \textbf{100.00} \\
                           & ResNet-56    & 97.14  & 69.19   & 62.79      & 98.44 & 94.31    & 99.13     & \textbf{100.00} & 99.94           & 99.64   & 67.74      & 98.44 & \textbf{100.00} & \textbf{100.00} & \textbf{100.00} \\
                           & MobileNetv2  & 98.99  & 79.46   & 82.32      & 98.74 & 97.10    & 99.58     & \textbf{99.97}  & 99.77           & 99.74   & 62.49      & 98.75 & \textbf{100.00} & \textbf{100.00} & \textbf{100.00} \\
                           & ShuffleNetv2 & 99.56  & 65.78   & 84.18      & 96.83 & 92.01    & 97.12     & \textbf{99.86}  & 99.89           & 98.80   & 58.13      & 96.83 & \textbf{100.00} & \textbf{100.00} & \textbf{100.00} \\ \midrule
\multirow{4}{*}{CIFAR-100} & VGG-19       & 92.93  & 64.37   & 90.48      & 95.12 & 88.31    & 91.58     & \textbf{98.17}  & 98.83           & 96.03   & 93.12      & 95.12 & 98.66           & 99.42           & \textbf{99.98}  \\
                           & ResNet-56    & 99.82  & 90.32   & 84.18      & 98.93 & 95.47    & 99.06     & \textbf{100.00} & 99.86           & 99.93   & 86.88      & 98.93 & 99.97           & \textbf{100.00} & \textbf{100.00} \\
                           & MobileNetv2  & 99.47  & 88.83   & 82.32      & 98.51 & 96.34    & 98.92     & \textbf{100.00} & 99.80           & 99.82   & 84.08      & 98.54 & 99.74           & \textbf{100.00} & \textbf{100.00} \\
                           & ShuffleNetv2 & 97.83  & 77.76   & 68.79      & 94.78 & 90.61    & 95.80     & \textbf{99.77}  & 99.84           & 98.99   & 71.25      & 94.78 & 99.79           & 99.96           & \textbf{100.00} \\ \midrule
\multirow{4}{*}{STL-10}    & VGG-19       & 89.25  & 77.70   & 42.97      & 95.12 & 95.07    & 83.65     & \textbf{99.00}  & 96.57           & 99.47   & 56.32      & 91.05 & 99.43           & 99.75           & \textbf{100.00} \\
                           & ResNet-56    & 96.78  & 73.94   & 28.12      & 91.41 & 90.34    & 96.90     & \textbf{99.80}  & \textbf{100.00} & 99.69   & 42.99      & 91.57 & \textbf{100.00} & \textbf{100.00} & \textbf{100.00} \\
                           & MobileNetv2  & \textbf{98.36}  & 77.32   & 33.59      & 91.99 & 96.66    & 91.95     & 97.10  & 99.85           & 99.40   & 64.24      & 89.65 & \textbf{100.00} & \textbf{100.00} & \textbf{100.00} \\
                           & ShuffleNetv2 & 96.82  & 79.08   & 33.13      & 90.82 & 94.75    & 92.17     & \textbf{98.50}  & \textbf{100.00} & 99.81   & 42.15      & 90.43 & \textbf{100.00} & 99.79           & \textbf{100.00} \\ \midrule
\multirow{4}{*}{ImageNet}  & VGG-19       & 81.25  & 67.09   & 42.97      & 95.12 & 95.07    & 93.48     & \textbf{99.52}  & 99.48           & 99.21   & 92.36      & 96.09 & \textbf{100.00} & 99.84           & \textbf{100.00} \\
                           & ResNet-152   & \textbf{96.78}  & 64.07   & 28.12      & 91.41 & 90.48    & 82.19     & 95.87  & 98.61           & 99.52   & 78.82      & 93.36 & 99.84           & 99.68           & \textbf{100.00} \\
                           & MobileNetv2  & 98.36  & 79.01   & 33.59      & 91.99 & 96.66    & 95.55     & \textbf{99.36}  & \textbf{100.00} & 99.84   & 83.16      & 92.38 & \textbf{100.00} & \textbf{100.00} & \textbf{100.00} \\
                           & ShuffleNetv2 & 96.82  & 79.01   & 33.13      & 90.82 & 94.75    & 89.19     & \textbf{99.05}  & 99.29           & 99.84   & 82.47      & 91.99 & \textbf{100.00} & 99.84           & \textbf{100.00} \\ \bottomrule
\end{tabular}
\end{table*}
\vspace{-10pt}

\begin{table*}[!htp]
\caption{\textbf{Attack success rate (ASR,\%) against robust models} on four benchmark datasets under query budget is set to \textbf{1000} and \textbf{10000}.}
\label{tab:robust}
\small
\centering
\renewcommand{\arraystretch}{1.1}
\setlength\tabcolsep{2.5pt}
\begin{tabular}{cccccccccccccccc}
\toprule
\multicolumn{2}{c}{Max$_Q$}                                   & \multicolumn{7}{c}{1000}                                                          & \multicolumn{7}{c}{10000}                                                    \\ 
\cmidrule(lr){1-2} \cmidrule(lr){3-9}  \cmidrule(lr){10-16}
Dataset                    & Model                            & Square & AdvFlow        & RS    & ASH            & NP    & Dif   & STBA           & Square & AdvFlow & RS    & ASH   & NP    & Dif             & STBA            \\ \midrule
\multirow{4}{*}{CIFAR-10}  & HYDRA                            & 13.33  & 36.24          & 29.87 & 28.99          & 7.49  & 21.89 & \textbf{67.30} & 20.86  & 58.60   & 31.56 & 29.18 & 19.68 & 47.19           & \textbf{98.24}  \\
                           & Wang$_{adv}$                     & 13.32  & 33.58          & 24.10 & 22.13          & 8.79  & 20.79 & \textbf{61.53} & 20.67  & 54.82   & 25.12 & 22.42 & 20.38 & 43.50           & \textbf{93.27}  \\
                           & Zhang$_{adv}$                    & 18.45  & 43.17          & 37.37 & 36.57          & 10.99 & 27.86 & \textbf{78.67} & 26.35  & 63.84   & 38.61 & 36.77 & 27.37 & 52.22           & \textbf{98.14}  \\
                           & FastAdv                          & 23.14  & 50.39          & 37.27 & 43.26          & 13.79 & 33.49 & \textbf{80.12} & 34.05  & 72.88   & 38.67 & 43.33 & 32.97 & 61.57           & \textbf{96.83}  \\ \midrule
\multirow{4}{*}{CIFAR-100} & Pang$_{adv}$                     & 26.38  & 38.35          & 51.23 & 40.25          & 13.39 & 30.47 & \textbf{82.17} & 33.89  & 61.38   & 52.82 & 40.25 & 27.07 & 56.41           & \textbf{98.34}  \\
                           & Efficient$_{adv}$                & 31.60  & 43.11          & 40.14 & 39.83          & 12.99 & 35.27 & \textbf{83.83} & 41.96  & 68.81   & 41.96 & 39.83 & 30.48 & \textbf{100.00} & 98.52           \\
                           & Proxy                            & 27.78  & 39.14          & 51.84 & 39.34          & 11.09 & 31.36 & \textbf{84.84} & 53.47  & 64.46   & 53.47 & 39.62 & 27.17 & \textbf{99.06}  & 98.87           \\
                           & Learn$_{adv}$                    & 28.16  & 36.42          & 53.88 & 48.21          & 9.99  & 32.69 & \textbf{95.93} & 55.03  & 67.78   & 55.03 & 49.12 & 28.97 & 87.76           & \textbf{99.58}  \\ \midrule
\multirow{4}{*}{STL-10}    & Free\_AT                         & 54.75  & 75.07          & 46.35 & \textbf{79.04} & 25.87 & 33.35 & 61.58          & 64.06  & 91.24   & 46.35 & 83.66 & 50.19 & 70.40           & \textbf{97.34}  \\
                           & FastAdv                          & 47.80  & \textbf{73.99} & 35.85 & 73.31          & 25.77 & 33.86 & 63.20          & 60.95  & 92.49   & 35.85 & 79.95 & 51.36 & 71.43           & \textbf{97.92}  \\
                           & TRADES                           & 60.52  & 69.49          & 24.78 & 80.21          & 51.25 & 54.75 & \textbf{90.93} & 82.01  & 94.93   & 24.78 & 85.09 & 92.22 & 93.57           & \textbf{100.00} \\
                           & MART                             & 53.23  & 71.73          & 35.85 & 73.24          & 48.55 & 49.69 & \textbf{88.97} & 72.59  & 92.60   & 29.61 & 80.08 & 82.88 & 87.02           & \textbf{100.00} \\ \midrule
\multirow{4}{*}{ImageNet}  & Slman\_${R18}$   & 39.26  & 31.12          & 36.33 & 49.02          & 36.33 & 8.16  & \textbf{70.02} & 52.73  & 70.78   & 75.35 & 67.38 & 59.53 & 70.78           & \textbf{97.53}  \\
                           & Slman\_${R50}$   & 17.97  & 34.91          & 18.55 & 36.13          & 18.55 & 8.35  & \textbf{54.84} & 33.79  & 70.97   & 64.58 & 55.08 & 43.97 & 70.97           & \textbf{95.45}  \\
                           & Engsrom\_${R18}$ & 18.36  & 36.81          & 18.55 & 40.82          & 18.55 & 9.49  & \textbf{44.97} & 38.67  & 73.06   & 65.62 & 62.11 & 46.69 & 73.06           & \textbf{95.64}  \\
                           & FastAdv                          & 36.72  & 51.80          & 30.86 & \textbf{55.86} & 30.86 & 18.41 & 50.47          & 55.47  & 86.15   & 76.39 & 72.27 & 53.70 & 86.15           & \textbf{96.20}  \\ \bottomrule
\end{tabular}
\end{table*}

\begin{figure*}[!ht]    
    \centering
        \setlength{\abovecaptionskip}{-5pt}
	\includegraphics[width=0.95\textwidth]{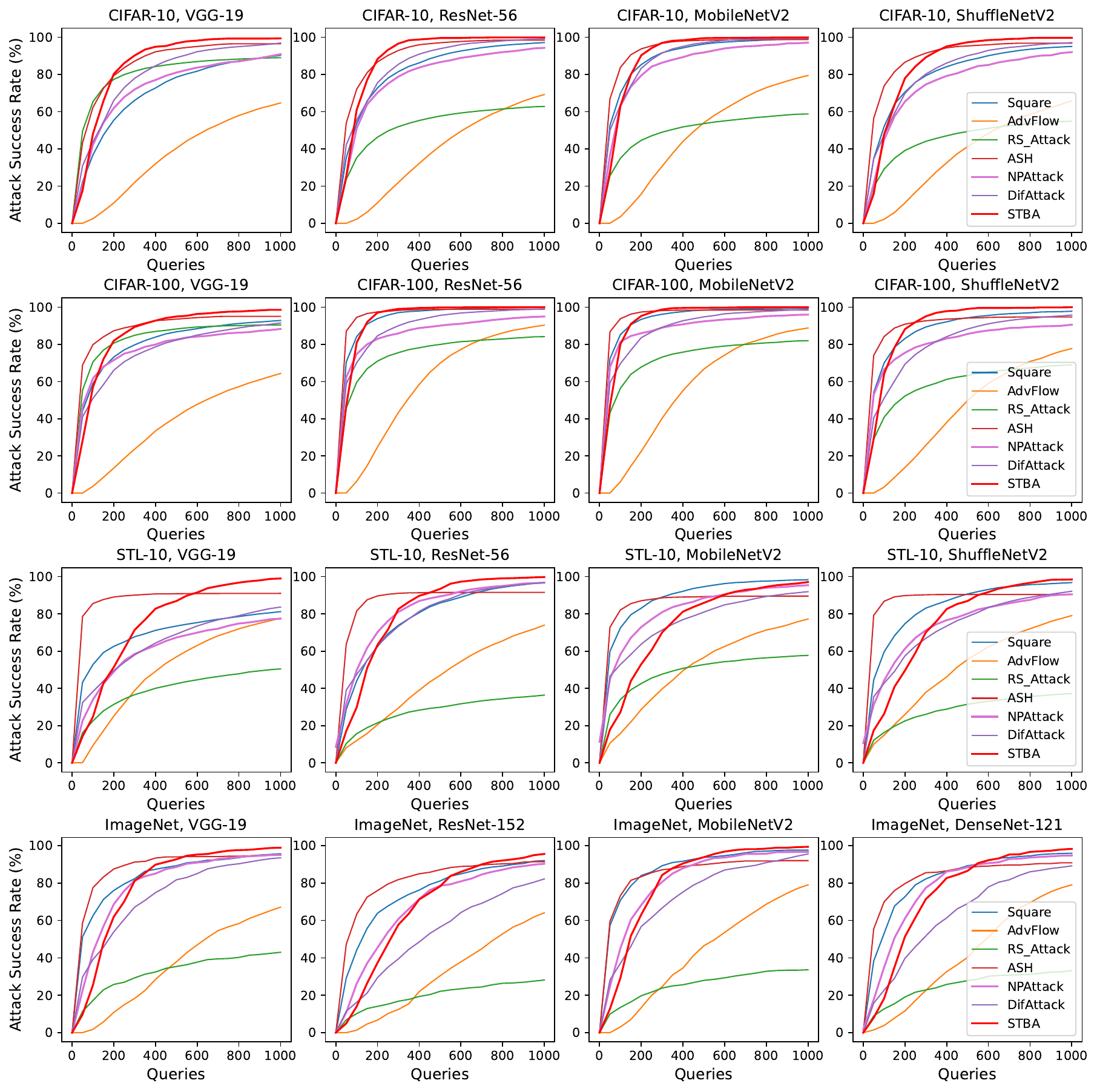}
	\caption{The attack success rate $vs.$ query numbers on \textbf{clean} models of the baselines and the proposed method, respectively, where the max query numbers are set to \textbf{1000}. The \textcolor{red}{\textbf{bold red}} lines are the results of the proposed STBA.}
	\label{fig:asr_clean_1k}
\vspace{-0.4cm}
\end{figure*}

\section{Experiments}
\label{Sec:experiments}
\subsection{Settings}
The maximum number of queries is set to 1000 and 10000 for STBA and the baselines to simulate a realistic attack scenario. And the experimental results of those methods are reproduced by the code released in the original paper with default settings. All the experiments are conducted on a GPU server with 4 * NVIDIA Tesla A100 40GB GPU, 2 * Xeon Glod 6112 CPU, and RAM 512GB. 

\textbf{Dataset:} We verify the performance of our method on {four} benchmark datasets for computer vision task, named CIFAR-10\footnote{http://www.cs.toronto.edu/~kriz/cifar.html} \cite{cifar-10}, CIFAR-100\footnote{http://www.cs.toronto.edu/~kriz/cifar.html} \cite{cifar-10}, STL-10\footnote{https://cs.stanford.edu/~acoates/stl10/} \cite{jmlr/CoatesNL11}, and ImageNet-1k\footnote{https://image-net.org/} \cite{cvpr/DengDSLL009}. In detail, CIFAR-10 contains 50,000 training images and 10,000 testing images with the size of 3x32x32 from 10 classes; CIFAR-100 has 100 classes, including the same number of training and testing images as the CIFAR-10; STL-10 has 10 classes, contains 500 training images and 800 testing images per class with image size of 3x96x96; ImageNet-1K has 1,000 categories, containing about 1.3M samples for training and 50,000 samples for validation. In particular, in this paper, we carry our attack on the whole images in testing datasets of CIFAR-10, CIFAR-100 and STL-10, in terms of ImageNet-1k, flowing previous works \cite{iclr/XieWZRY18,cvpr/DongLPS0HL18,compsec/PengLWCWCZ21,icics/LiuZLZWZ23}, we are using its subset datasets from NIPS2017 Adversarial Learning Challenge\footnote{https://www.kaggle.com/competitions/nips-2017-non-targeted-adversarial-attack/data}.

\textbf{Victim Models:} For normal trained models, we adopted the pre-trained VGG-19\_bn (VGG-19) \cite{corr/SimonyanZ14a}, ResNet-56 \cite{cvpr/HeZRS16}, MobileNet-V2 \cite{cvpr/SandlerHZZC18} and ShuffleNet-V2 \cite{eccv/MaZZS18} for CIFAR-10 and CIFAR-100, all the models' weights are provided in the GitHub Repository \footnote{https://github.com/chenyaofo/pytorch-cifar-models}, and we train these models on STL-10 by ourselves. For ImageNet, we use the PyTorch officially pre-trained model VGG-19 \cite{corr/SimonyanZ14a}, ResNet-152 \cite{cvpr/HeZRS16}, MobileNet-V2 \cite{cvpr/SandlerHZZC18} and DenseNet-121 \cite{cvpr/HuangLMW17} as the victim models, respectively. 

To investigate the performance of the proposed method in attacking robust models, we select some of the most recent defense techniques as follows: they are including HYDRA \cite{nips/Sehwag0MJ20}, Wang$_{adv}$ \cite{iclr/0001ZY0MG20}, Zhang$_{adv}$ \cite{icml/ZhangYJXGJ19}, FastAdv \cite{iclr/WongRK20} for CIFAR-10, Pang$_{adv}$ \cite{icml/PangLYZY22}, Efficient$_{adv}$ \cite{nips/AddepalliJR22}, Proxy \cite{iclr/SehwagMHD0CM22}, Learn$_{adv}$ \cite{iccv/Cui0WJ21} for CIFAR-100, FreeAT \cite{nips/ShafahiNG0DSDTG19}, FastAdv \cite{iclr/WongRK20}, TRADES \cite{icml/ZhangYJXGJ19} and MART \cite{iclr/0001ZY0MG20} for STL-10, and Salman$_{R50}$ \cite{nips/SalmanIEKM20}, Engstrom$_{adv}$ \cite{engstrom2019adversarial}, FastAdv \cite{iclr/WongRK20}, Salman$_{R18}$ \cite{nips/SalmanIEKM20} for ImageNet. For all the models of CIFAR-10, CIFAR-100, and ImageNet, we use their implementation in the robustbench toolbox\footnote{https://github.com/RobustBench/robustbench} \cite{nips/CroceASDFCM021} and the models' weights are also provided in this toolbox \cite{nips/CroceASDFCM021}. For all these models, we chose their $L_\infty$-norm version parameters because we carry out $ L_\infty $-norm attacks in this paper. Similar to normal models, we train the robust models of STL-10 by ourselves with the help of their officially provided codes, and we listed the recognition performance of all these aforementioned models on tested images in the Table. \ref{tab:victim}.

Furthermore, three online vision models are involved in evaluating the attack in the physical world, including Google Cloud Vision\footnote{https://cloud.google.com/vision}, Tencent Cloud\footnote{https://cloud.tencent.com/} and Baidu AI Cloud\footnote{https://cloud.baidu.com/}. 

\begin{figure*}[!ht]    
    \centering
	\includegraphics[width=0.95\textwidth]{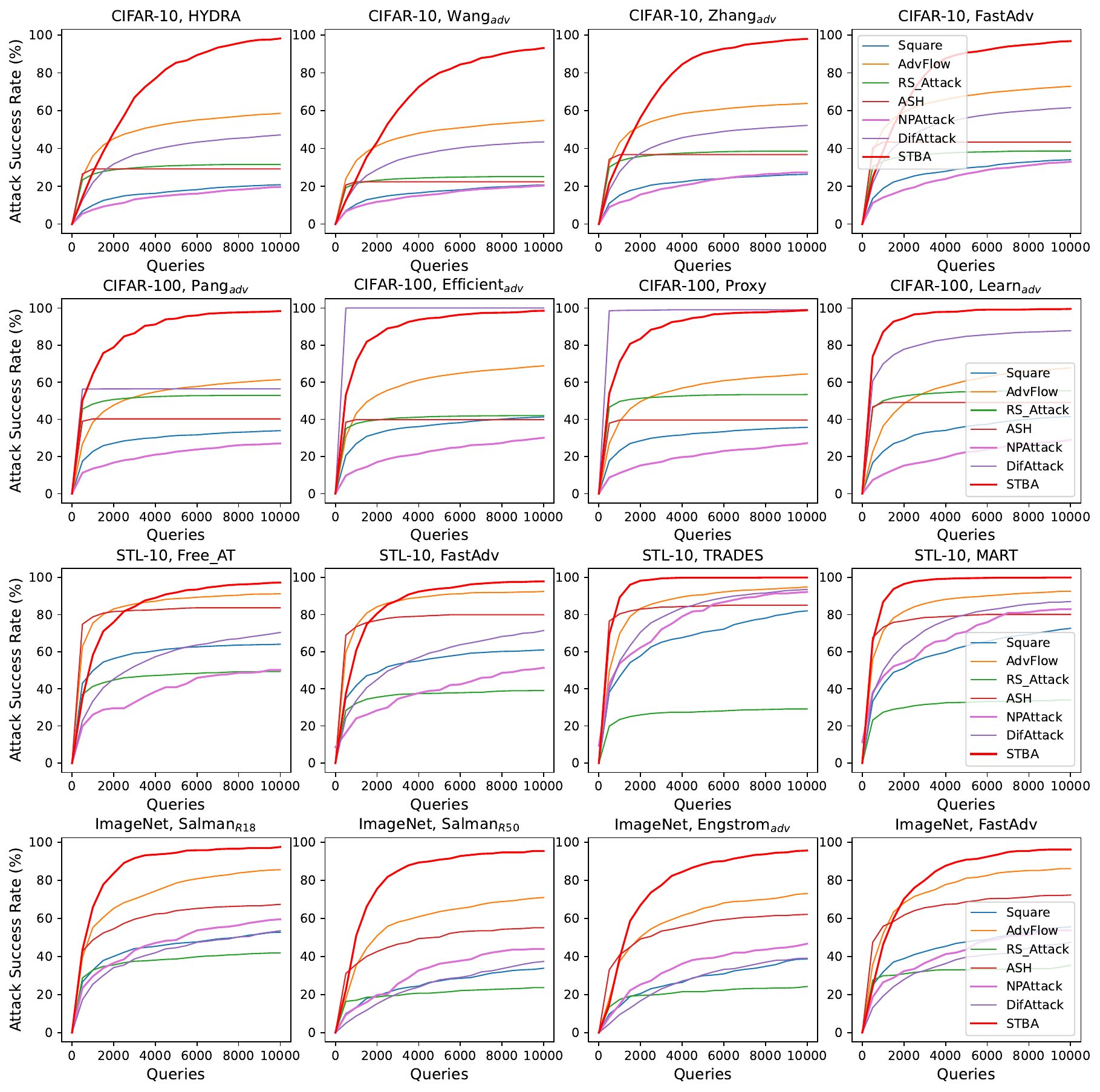}
     \setlength{\abovecaptionskip}{-5pt}
	\caption{The attack success rate $vs.$ query numbers on \textbf{robust} models of the baselines and the proposed method, respectively, where the max query numbers are set to \textbf{10000}. The \textcolor{red}{\textbf{bold red}} lines are the results of the proposed STBA.}
	\label{fig:asr_robust_10k}
 \vspace{-0.4cm}
\end{figure*}

\textbf{Baselines:} The baselines for query-limited score-based black-box attack settings are Square Attack \cite{eccv/AndriushchenkoC20}, AdvFlow \cite{nips/DolatabadiEL20}, RS\_Attack \cite{aaai/CroceASF022}, ASH \cite{mir/LiH00S22}, NPAttack \cite{pr/BaiWZJX23} and DifAttack \cite{aaai/00710ZT24}. For Square Attack, AdvFlow, ASH, and NPAttack, we set the $L_\infty=0.05$, and for RS\_Attack, we set the patch size as 3x3 for CIFAR-10 and CIFAR-100, 8x8 for STL-10, and 20x20 for ImageNet. For AdvFlow, we use the officially pre-trained flow weights on CIFAR-10 and train on other datasets by ourselves, and for the DifAttack, we use the SqueezeNet \cite{corr/IandolaMAHDK16} as the surrogate model.

\textbf{Metrics:} We compare our proposed method with several query-efficient black-box attack techniques concerned with Attack Success Rate (ASR), Average Query number (Average.Q) and Median Query number (Med.Q), and image quality involving LPIPS \cite{cvpr/ZhangIESW18}, DISTS \cite{pami/DingMWS22}, MAD \cite{jei/LarsonC10}, BRISQUE \cite{tip/MittalMB12}, FID \cite{nips/HeuselRUNH17}, SSIM \cite{tip/WangBSS04}, PSNR, VIF \cite{HAN2013127} and FSIM \cite{zhang2011fsim}. 

\begin{figure*}[!ht]    
    \centering
	\includegraphics[width=0.90\textwidth]{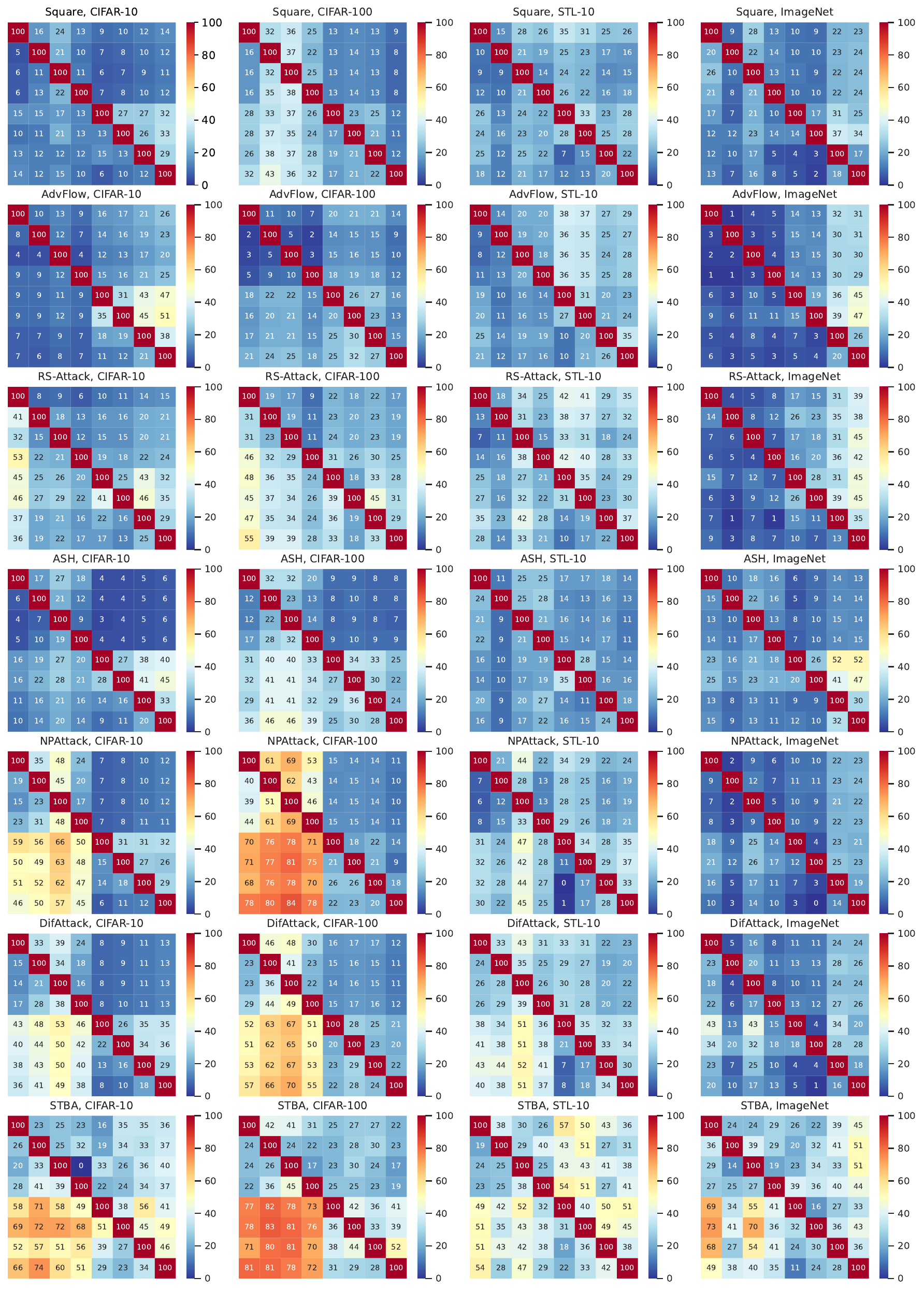}
    \setlength{\abovecaptionskip}{-5pt} 
	\caption{Confusion matrix of transferability for adversarial attacks generated by baselines and the proposed STBA on CIFAR-10, CIFAR-100, STL-10 and ImageNet, respectively. The row represents which model is targeted during the generation of adversarial examples, while the column represents which model is attacked by such synthesized examples.}
	\label{fig:confusion}
 \vspace{-0.4cm}
\end{figure*}

\subsection{Quantitative Comparison with the Query-Efficient Black-box Attacks}
In this subsection, we will evaluate the proposed STBA and the baselines in attacking performance and query efficiency on the validation dataset of CIFAR-10, CIFAR-100, STL-10 and the whole NIPS2017 dataset. Note that, in this part, we use RS, NP, and Dif to represent RS\_Attack, NPAttack, and DifAttack, respectively.

Table. \ref{tab:clean} and Table. \ref{tab:robust} show the attack results of all attacks on the normally trained and robustly trained models, respectively. As can be seen, STBA can improve upon the performance of baseline methods in all of the target models in terms of the attack success rate. For normally trained models, even when the max query is set to 1000, STBA can obtain nearly 100\% attack success rate, while the baselines are struggling to achieve such a high attack success rate, in most cases, their attack success rates are lower than 95\% in this query-limited situation. Besides, when attacking robust models, baseline methods perform poorly under the max query is limited to 1000, while STBA can obtain 61.53 $\sim$ 80.12\%, 82.17 $\sim$ 95.93\%, 61.58 $\sim$ 90.93\% and 44.97 $\sim$ 70.02\% attack success rate on the four different benchmark datasets, respectively. Even if we increased the max query budget to 10000, existing methods would still find it hard to attack the robust models successfully and can only get 19.68 $\sim$ 92.22\% attack success rate. In contrast, STBA can obtain 93.27 $\sim$ 100\% attack success rate.

The empirical results indicated that although previous techniques show fantastic performance in query-based black-box attack settings, once we put a relatively extreme limit on the maximum number of queries, these methods will lose their advantages and show {dissatisfactory} results. Besides, such results also verify our assumption that the previous techniques struggle to attack the robustly trained model successfully.

Furthermore, we exhibit the attack performance and query times of all methods in Fig. \ref{fig:asr_clean_1k} and Fig. \ref{fig:asr_robust_10k}, respectively. From Fig. \ref{fig:asr_robust_10k}, attacking results on robust models, we can clearly find that under a limited query budget, i.e., 5000, the baselines can hardly achieve 40+\% attack success rate on the CIFAR-10 dataset, while our method can obtain nearly 80\% attack success rate. For the ImageNet dataset, our method can achieve a more satisfactory attack performance as the query budget grows and wins the best results with the maximal query budget. In contrast, others can only obtain about a 60\% attack success rate and cannot even obtain a higher attack success rate by increasing the query budget. These results fully support our claims again: 1) Although existing black-box attacks can achieve high attack success rates, even reaching 100\% attack success rate in some situations, they require a large amount of interaction from the victim model to obtain useful information to optimize perturbations. Once the number of queries is limited, their attack performance will be significantly affected. 2) Although existing black-box attack methods have achieved satisfactory attack results on normally trained black-box models, they are still difficult to successfully attack robust models with a high attack success rate, whose robustness is usually improved by adversarial training. This dramatically limits their attack effect in practical situations because most of the deep learning models deployed in the real world are relatively robust or equipment defense mechanisms. 3) Existing adversarial training-based robust models still show vulnerability against the newly proposed attacks, which are not involved in providing adversarial examples in the process of adversarial training or fine-tuning.

\begin{figure*}[ht]
    \centering
    \setlength{\abovecaptionskip}{-5pt}
    \includegraphics[width=0.8\textwidth]{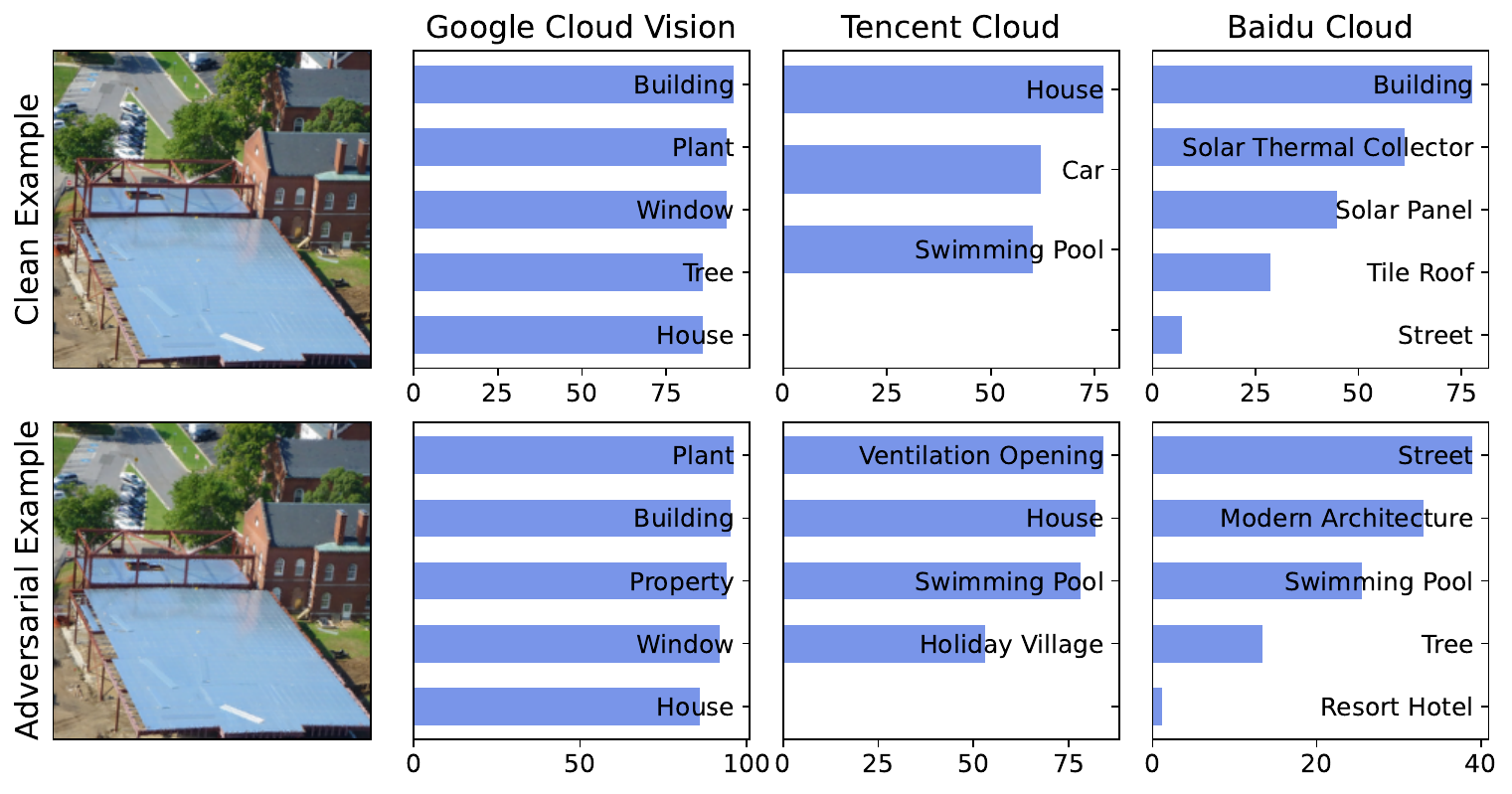}
    \caption{The top-5 prediction confidence of the clean image and its corresponding adversarial counterpart returned by Google Cloud Vision, Tencent Cloud and Baidu Cloud, respectively.}
    \label{fig:top-5}
    \vspace{-10pt}
\end{figure*}

\begin{figure}[htp]
    \centering
    \setlength{\abovecaptionskip}{-2pt}
    \includegraphics[width=0.49\textwidth]{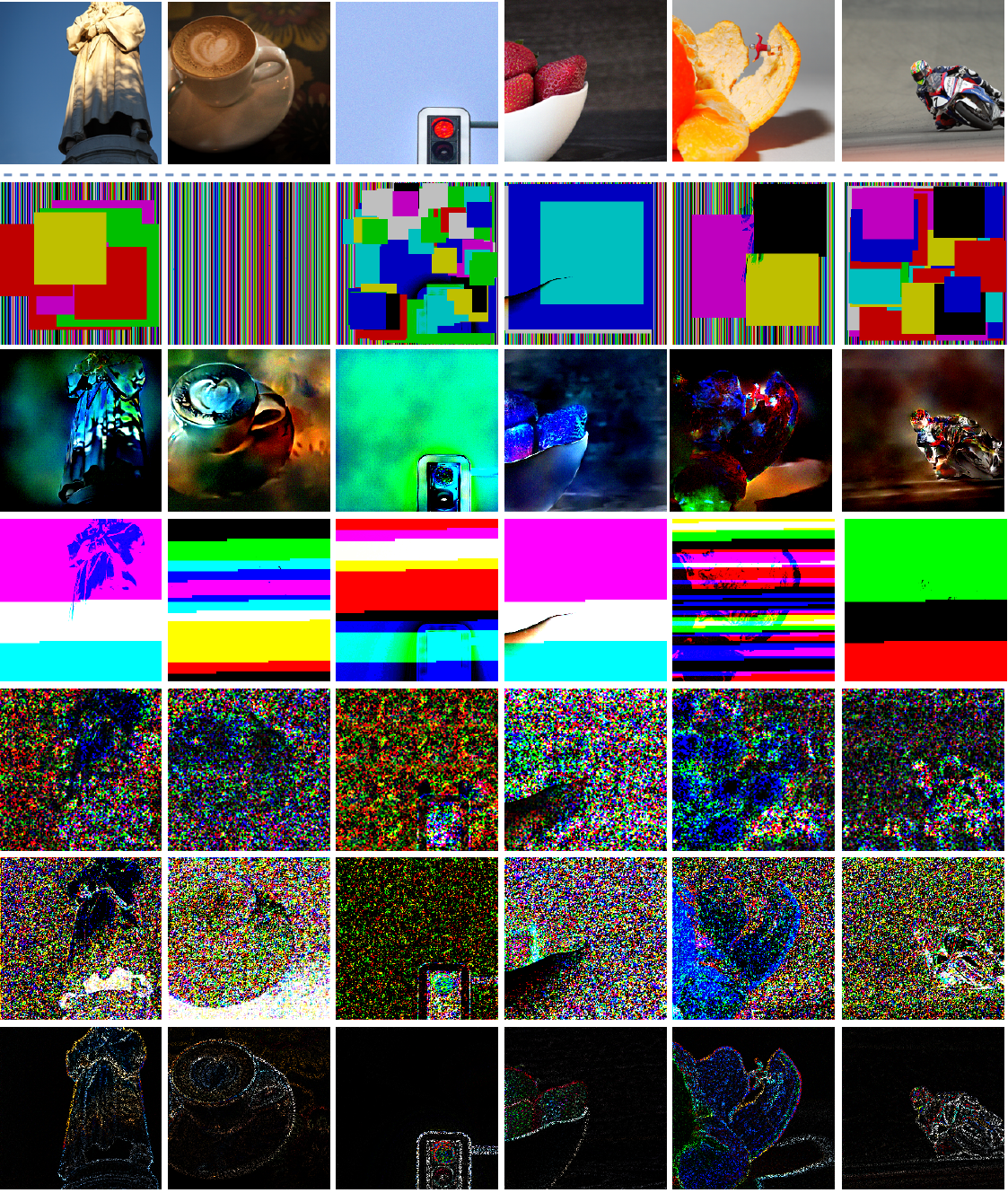}
    \caption{The clean example and their corresponding adversarial noise generated by baselines and the proposed STBA. The first row is the clean images, and the flowing rows are the adversarial noise generated by Square Attack \cite{eccv/AndriushchenkoC20}, AdvFlow \cite{nips/DolatabadiEL20}, ASH \cite{mir/LiH00S22}, NPAttack \cite{pr/BaiWZJX23}, DifAttack \cite{aaai/00710ZT24}, and STBA, respectively.}
    \label{fig:noise}
    \vspace{-10pt}
\end{figure}

\subsection{Transferability}
\label{transferability}
The motivation of the current work states that the generation of adversarial examples is generally based on the assumption of transferability, which means that an adversarial example generated according to a model can be used to attack other unseen models. To see if this assumption is valid in black-box settings, here, we follow the previous works \cite{nips/DolatabadiEL20,aaai/ZhaoCWL20} to examine the transferability of the generated adversarial examples across different models on CIFAR-10, CIFAR-100, STL-10 and ImageNet. Specifically, we generate adversarial examples for the aforementioned eight models shown in Table \ref{tab:victim} for each dataset, including four normal-trained and four adversarial-trained. We randomly selected 1000 images from the CIFAR-10, CIFAR-10, and STL-10 test set, and selected all images from ImageNet (NIPS2017), which are classified correctly by these models, whereas the corresponding adversarial examples are misclassified, for this experiment. Where such generated adversarial examples are used to attack the other models. Here, we set the maximal query number to 1000 for all baselines.

We illustrate the results of this part with the ASR confusion matrix on these four datasets, as shown in Fig. \ref{fig:confusion}. The row represents which model is targeted during the generation of adversarial examples, while the column represents which model is attacked by the generated examples. From this figure, we can see that the highest transferability ASR on four datasets of STBA can achieve 74\%, 83\%, 57\%, and 73\%, respectively. In contrast, the comparable baseline method NPAttack is 66\%, 84\%, 47\%, and 26\%, respectively. The rest of the baselines even exhibit poor transferability. It means that the examples generated by STBA produce higher transfer ASR in most cases than baselines, especially for the STL-10 and ImageNet datasets, which validate the superior transferability of STBA. Meanwhile, this phenomenon shows that baselines heavily rely on the feedback of the target model during each query and cannot extract transferable features. By contrast, our method learns the adversarial flow field $\bm{f}$ that does not collapse to a specific model.

In addition, we were intrigued to discover that AEs generated on robust models exhibit superior transfer attack performance. This finding suggests that researchers can effectively leverage AEs crafted on robust models to conduct transfer attacks on various other models. Such a strategy allows for a more efficient exploration of vulnerabilities across existing models. Furthermore, this approach can significantly aid in the identification of critical weaknesses, thereby encouraging defenders to develop more robust and effective defense mechanisms. This could lead to enhancements in overall model robustness, promoting more secure AI applications.

\subsection{Visual Quality}
Different from the previous works using $L_p$-norm to evaluate the generated adversarial images' quality, in this paper, we follow previous work \cite{mm/AydinSKHT21} and use the following perceptual metrics to evaluate the adversarial examples generated by our method and baselines, such as: Learned Perceptual Image Patch Similarity (LPIPS) metric \cite{cvpr/ZhangIESW18}, Deep Image Structure and Texture Similarity (DISTS) index \cite{pami/DingMWS22}, Mean Absolute Difference (MAD) \cite{jei/LarsonC10}, Blind/Referenceless Image Spatial Quality Evaluator (BRISQUE) \cite{tip/MittalMB12}, Fréchet Inception Distance (FID) \cite{nips/HeuselRUNH17}, Structure Similarity Index Measure (SSIM) \cite{tip/WangBSS04}, Peak Signal-to-Noise Ratio (PSNR), Visual Information Fidelity (VIF) \cite{HAN2013127} and Feature Similarity Index Measure (FSIM) \cite{zhang2011fsim} to assess the generated images' qualities in terms of various aspects. The image quality assessment toolkit we used in the experiments of this part is IQA\_pytorch\footnote{https://www.cnpython.com/pypi/iqa-pytorch}. Note that we omitted RS\_Attack in this part because it is a patch-based attack.

The quantitative experimental results of images' quality can be seen in Table. \ref{tab:visual}, which indicated that the proposed method has the lowest MAD and BRISQUE (the lower is better), are 3.0883 and 16.4823, respectively, and has the highest SSIM, PSNR, VIF and FSIM (the higher is better), achieving 0.9879, 37.3324, 0.8258 and 0.9873, respectively, in comparison to the baselines on ImageNet dataset. The results show that the proposed method is superior to the existing attack methods.

\begin{table}[]

\caption{\textbf{Visual quality evaluation results} calculated on adversarial examples crafted by Square, AdvFlow, ASH, NPAttack, DifAttack, and the proposed STBA.}
\label{tab:visual}
\small
\centering
\renewcommand{\arraystretch}{1.1}
\setlength\tabcolsep{1pt}
\begin{tabular}{ccccccc}
\toprule
Metric               & Square  & AdvFlow                                & ASH     & NPAttack & DifAttack & STBA             \\ \midrule
LPIPS $\downarrow$   & 0.2448  & \textbf{0.0135} & 0.1298  & 0.1168   & 0.2079    & 0.0162           \\
DISTS $\downarrow$   & 0.2677  &  \textbf{0.0366} & 0.1625  & 0.1340   & 0.1869    & 0.0680           \\
MAD $\downarrow$     & 90.7684 & 29.4512                                & 73.7850 & 66.6841  & 87.9537   & \textbf{3.0883}  \\
BRISQUE $\downarrow$ & 29.7395 & 17.0994                                & 20.3779 & 17.2939  & 26.6368   & \textbf{16.4823} \\
FID $\downarrow$     & 97.0623 & \textbf{8.5572} & 50.4029 & 29.9010  & 44.4464   & 13.9358          \\
SSIM $\uparrow$      & 0.8202  & 0.9865                                 & 0.9516  & 0.9122   & 0.7969    & \textbf{0.9879}  \\
PSNR $\uparrow$      & 26.1883 & 33.3259                                & 26.1909 & 31.3421  & 28.1149   & \textbf{37.3324} \\
VIF $\uparrow$       & 0.6240  & 0.8137                                 & 0.7887  & 0.5404   & 0.4456    & \textbf{0.8258}  \\
FSIM $\uparrow$      & 0.9025  & 0.9721                                 & 0.9630  & 0.9410   & 0.9001    & \textbf{0.9873}  \\ \bottomrule
\end{tabular}

\end{table}

Furthermore, to better observe the difference between the adversarial examples generated by our method and the baselines from the visual aspect, we draw the adversarial perturbation generated on ImageNet by baselines and the proposed method in Fig. \ref{fig:noise}, the target model is pre-trained VGG-16. The first row is the benign examples, and the following are the adversarial noise of Square \cite{eccv/AndriushchenkoC20}, AdvFlow \cite{nips/DolatabadiEL20}, ASH \cite{mir/LiH00S22}, NPAttack \cite{pr/BaiWZJX23}, DifAttack \cite{aaai/00710ZT24} and our method, respectively. Noted that, for better observation, we magnified the noise by multiplying a factor of 25. From Fig. \ref{fig:noise}, we can clearly observe that baseline methods distort the image without ordering. In contrast, the adversarial perturbations generated by our method are focused on the target object instead of disturbing the whole image, and its noise contains more semantic information and is more imperceptible to human eyes.

\subsection{Performance on Physical Online Vision Models}
To evaluate the adversarial examples' effectiveness in real-world scenarios, we conduct experiments of attacking the online models on Google Cloud Vision, Tencent Cloud and Baidu AI Cloud, whose structures and parameters and datasets used for training in their platforms are entirely sightless for us. We use the ImageNet dataset to carry out our attack on the pre-trained ResNet-152 model and then we randomly choose an adversarial image and its clean counterpart to validate on three online image recognition platforms. We draw the returned top-5 results in Fig. \ref{fig:top-5}, which shows that the attack has successfully changed the image's top-5 prediction results to a large degree \footnote{We call these APIs on March 24, 2024.}. Taking Baidu AI Cloud for instance, its top-5 outputs of the clean image like Building (77.67\%), Solar Thermal Collector (61.18\%), Solar Panel (44.82\%), Tile Roof (28.49\%), Street (7.09\%). However, after attacking, its top-5 predictions changed to Street (38.97\%), Modern Architecture (32.94\%), Swimming Pool (25.50\%), Tree (13.43\%) and Resort Hotel (1.17\%). Besides, for Tencent Cloud, the adversarial example even increases the output labels from 3 to 4. This phenomenon indicates that our attack method is also applicable to the real physical world.

\subsection{Ablation Study}
\label{sec:ablation}

\begin{table}[]
\caption{The attack performance, query efficiency, and visual quality of applying the spatial transform to \textbf{High- $vs.$ Low-frequency} part.}
\label{tab:abation_high_low}
\small
\centering
\setlength\tabcolsep{4pt}
\begin{tabular}{ccccccc}
\toprule
Dataset                    & Frequency & ASR   & Avg.Q  & Med.Q  & LPIPS & DISTS \\ \midrule
\multirow{2}{*}{CIAFAR-10} & High & 99.40 & 142.47 & 111.00 & 0.02  & 0.05  \\
                           & Low  & 31.40 & 387.19 & 342.00 & 0.05  & 0.07  \\ \midrule
\multirow{2}{*}{CIFAR-100} & High & 99.60 & 132.58 & 89.00  & 0.02  & 0.05  \\
                           & Low  & 48.10 & 363.68 & 298.00 & 0.06  & 0.08  \\ \midrule
\multirow{2}{*}{STL-10}    & High & 99.10 & 241.65 & 190.00 & 0.02  & 0.06  \\
                           & Low  & 43.40 & 385.62 & 337.00 & 0.05  & 0.10  \\ \midrule
\multirow{2}{*}{ImageNet}  & High & 99.50 & 205.34 & 169.00 & 0.02  & 0.06  \\
                           & Low  & 27.70 & 293.26 & 169.00 & 0.06  & 0.09  \\ \bottomrule
\end{tabular}
\end{table}

\begin{figure}[htp]
    \centering
    \setlength{\abovecaptionskip}{-5pt}
    \includegraphics[width=0.49\textwidth]{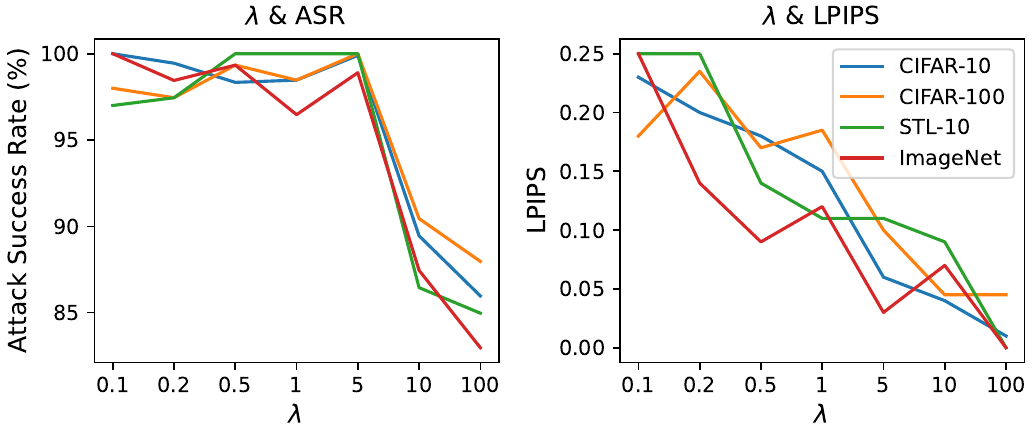}
    \caption{Ablation study of attack performance and generated adversarial examples' quality under different loss weighting factor $\lambda$.}
    \label{fig:lambda}
    \vspace{-0.2cm}
\end{figure}

\begin{figure}[htp]
    \centering
    \setlength{\abovecaptionskip}{-5pt}
    \includegraphics[width=0.49\textwidth]{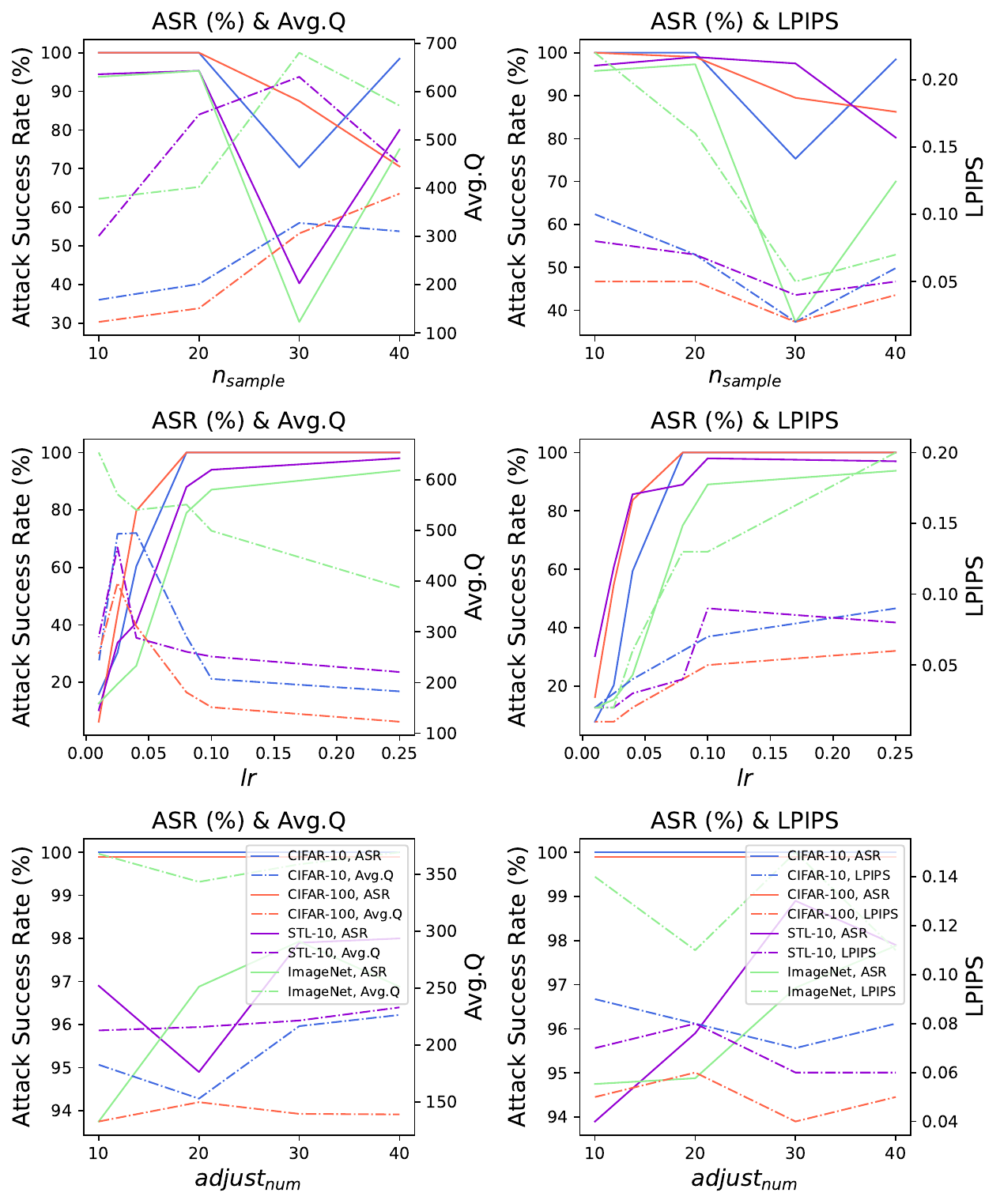}
    \caption{Ablation study of attack performance and generated adversarial examples' visual quality under different hyper-parameters. The critical hyper-parameters involved here are the number of samples $n_{sample}$, the learning rate $lr$ and the number of adjustments of flow field budget $\xi$ $adjust_{num}$. We show the results of attack success rate, average query number and image quality in each row, respectively. The lines with different colors represent the result on different datasets, the solid lines represent the attack success rate, and the dashed lines indicate the average query number (Avg.Q) and image quality, respectively.}
    \label{fig:ablation}
    \vspace{-0.2cm}
\end{figure}

In this section, we extend the ablation studies to verify the attack performance (including attack success rate and query times) and the generated adversarial examples' visual quality under different hyper-parameter settings.

We first investigate the effectiveness of the proposed strategy in terms of applying the spatial transform to the high-frequency part to improve the attack performance and the visual quality of generated adversarial examples. We list the results in Table \ref{tab:abation_high_low}, which verified our claim and shows that applying spatial transform to the high-frequency part not only gets a good attack performance but also improves the query efficiency and image quality. The test images are randomly selected and correctly classified 1000 images for CIFAR-10, CIFAR-100, and STL-10, while the ImageNet (NIPS) are all correctly classified images. Here, the maximal query budget is set to 1000, and the victim model is VGG-19.

Besides, we also investigate the influence of loss function balancing factor $\lambda$ on the performance of the proposed method. Specifically, we conduct a dissection experiment to systematically vary the balancing factor $\lambda$ within the loss function and observe its effects on attack success rate and the image quality of the generated adversarial examples. The results Fig. \ref{fig:lambda} show that the attack can maintain a success rate when the $\lambda$ is set to a small value, i.e., 0.1 to 5, and has an obvious drop when $\lambda$ is greater than 5. While the image quality decreases with the $\lambda$ increase. Therefore, to balance the attack performance and image quality of the generated images, we set the $\lambda = 5$ in this paper according to these empirical results.

Finally, in our method, other involved critical hyper-parameters are as follows: the number of samples $n_{sample}$ to estimate the gradient, the learning rate $lr$ and the number of adjustments of flow field budget $\xi$, $adjust_{num}$. First of all, we set $\sigma=2*lr$ and do the experiments under a predefined hyper-parameters set; we illustrate the results in Fig. \ref{fig:ablation}. Specifically, we set the $n_{sample} \in \{ 10,20,30,40\}$, $lr \in \{ 0.25,0.1,0.08,0.04,0.025,0.01\}$ and $adjust_{num} \in \{ 10,20,30,40\}$ and extent attacks on the ResNet (ResNet-56 for CIFAR and STL-10 dataset, and ResNet-152 for ImageNet dataset, respectively.) under the max query number is set to 1000. As the results show, for $n_{sample}$, we set $n_{sample}=10$ to get the highest attack success rate and lowest average query number and set $n_{sample}=30$ to get the best image quality, but the attack success rate will be faded for most victim datasets. For $lr$, we set $lr=0.1$ to balance the attack success rate, average query number, and image quality. For $adjust_{num}$, STBA achieves the best attack performance and image quality when $adjust_{num}=30$. Therefore, in our experiments, to obtain the adversarial images with high attack success and image quality, we set the most essential hyper-parameters as $n_{sample}=10$, $lr=0.1$ and $adjust_{num}=20$, respectively.

\section{Conclusions and Limitations}
\label{Sec:conclusion}
To evaluate the vulnerability and robustness of existing normally trained and robustly trained DNN models under query-limited black-box situations, in this paper, we present a novel non-noise additional method, called STBA, to build adversarial examples. The proposed STBA combines performing the spatial transformation to the high-frequency part of the image and gradient estimation to search for the best flow field to synthesize adversarial examples. Empirically, the adversarial examples generated by STBA are imperceptible to the human eyes and significantly increase query efficiency; it only needs dozens and hundreds of queries to obtain a practical adversarial example in attacking normally trained and robustly trained models, respectively. Extensive experiments show that the proposed method is superior to the existing methods in terms of attack success rate, query times and imperceptibility. Even on robust models, it can obtain 93.27 $\sim$ 100\% attack success rate on benchmark datasets. Benefitting from generating adversarial examples without noise-adding, the proposed STBA provides a new efficient way to evaluate the vulnerability of existing classifiers and further enhance their robustness performance using techniques like fine-tuning or adversarial training. 

Although our method can achieve better attack results and visual imperceptibility under query-limited black-box attack settings, we still cannot guarantee a 100\% attack success rate all the time with such minor modifications. Besides, we find that the modification in some generated adversarial examples exhibits a jagged distortion because we struggle to calculate a more elaborate transform matrix with gradient estimation in a black-box attack setting. Further, we will solve this limitation by extending spatial transform operations in other image spaces, such as YUV, LAB, et al., instead of RGB space to generate more imperceptible adversarial examples in black-box scenarios under guaranteed attack performance.

\section*{Acknowledgments}
Any opinions, findings and conclusions or recommendations expressed in this material are those of the author(s) and do not reflect the views of the National Research Foundation, Singapore and Infocomm Media Development Authority, Yunnan Province expert workstations, Yunnan Fundamental Research Projects.

\bibliographystyle{IEEEtran}
\bibliography{ref}


\section*{Biography Section}

\begin{IEEEbiography}[{\includegraphics[width=1in,height=1.25in,clip,keepaspectratio]{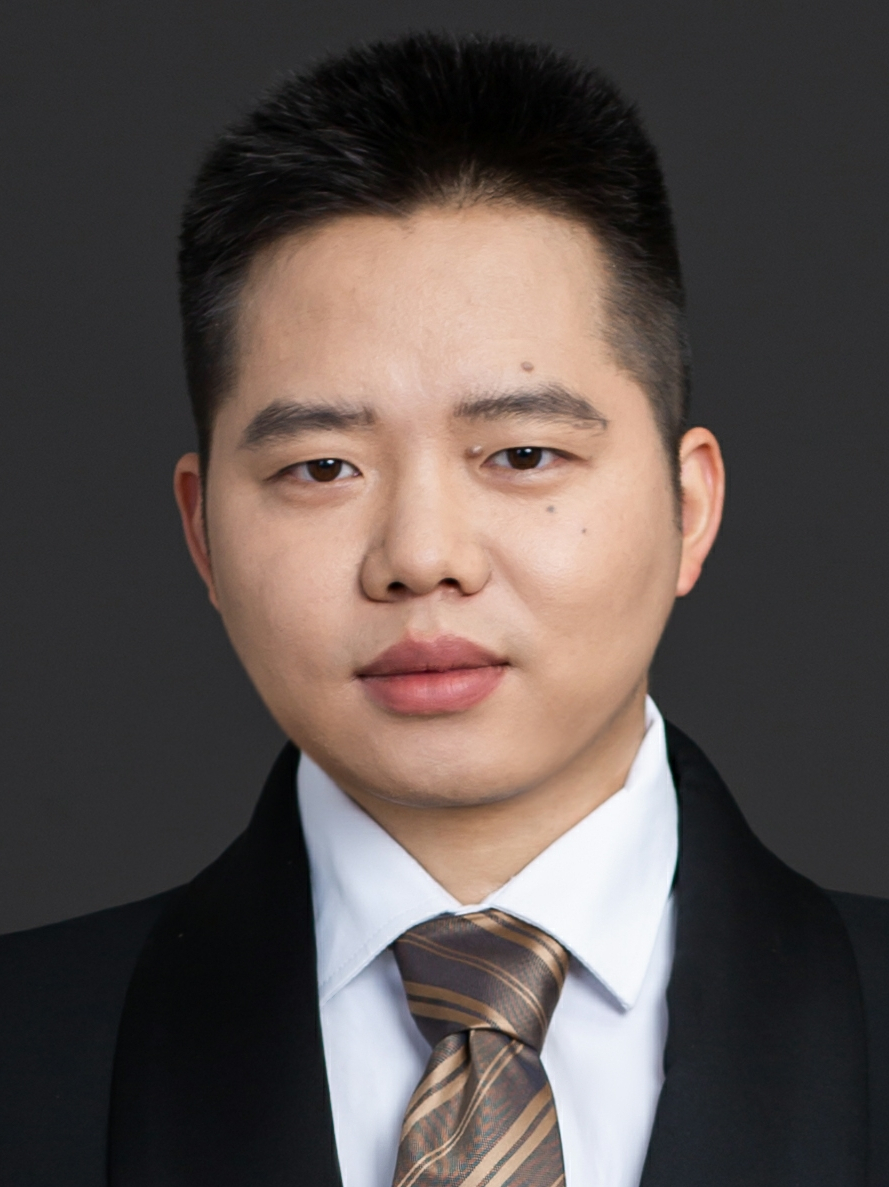}}]{Renyang Liu} (Graduate Student Member, IEEE) earned his B.E. in Computer Science from Northwest Normal University in 2017 and his Ph.D. from Yunnan University in 2024. He served as a Joint-training Ph.D. student at Nanyang Technological University's College of Computing and Data Science from 2022 to 2023. Currently, he is a Research Fellow at the Institute of Data Science, National University of Singapore. His research primarily focuses on security issues of multimodal models, with interests spanning AI security, machine unlearning, data privacy, and computer vision.
\end{IEEEbiography}

\vspace{-10pt}

\begin{IEEEbiography}[{\includegraphics[width=1in,height=1.25in,clip,keepaspectratio]{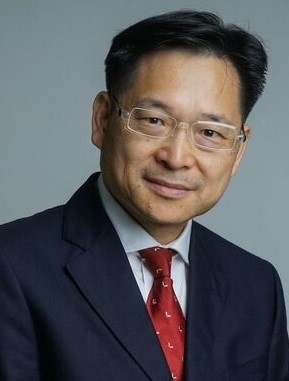}}]{Kwok-Yan Lam} (Senior Member, IEEE) received his B.Sc. degree (1st Class Hons.) from University of London, in 1987, and Ph.D. degree from University of Cambridge, in 1990. He is the Associate Vice President (Strategy and Partnerships) and Professor in the College of Computing and Data Science at the Nanyang Technological University, Singapore. He is currently also the Director of the Strategic Centre for Research in Privacy-Preserving Technologies and Systems (SCRiPTS). From August 2020, he is on part-time secondment to the INTERPOL as a Consultant at Cyber and New Technology Innovation. Prior to joining NTU, he has been a Professor of the Tsinghua University, PR China (2002–2010) and a faculty member of the National University of Singapore and the University of London since 1990. He was a Visiting Scientist at the Isaac Newton Institute, Cambridge University, and a Visiting Professor at the European Institute for Systems Security. In 1998, he received the Singapore Foundation Award from the Japanese Chamber of Commerce and Industry in recognition of his research and development achievement in information security in Singapore. His research interests include Distributed Systems, Intelligent Systems, IoT Security, Distributed Protocols for Blockchain, Homeland Security and Cybersecurity.
\end{IEEEbiography}

\vspace{-10pt}

\begin{IEEEbiography}[{\includegraphics[width=1in,height=1.25in,clip,keepaspectratio]{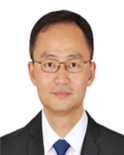}}]{Wei Zhou}
 received the Ph.D. degree from the University of Chinese Academy of Sciences. He is currently a Full Professor with the Software School, Yunnan University. His current research interests include the distributed data intensive computing and bio-informatics. He is currently a Fellow of the China Communications Society, a member of the Yunnan Communications Institute, and a member of the Bioinformatics Group of the Chinese Computer Society. He won the Wu Daguan Outstanding Teacher Award of Yunnan University in 2016, and was selected into the Youth Talent Program of Yunnan University in 2017. Hosted a number of National Natural Science Foundation projects.
\end{IEEEbiography}

\vspace{-10pt}

\begin{IEEEbiography}[{\includegraphics[width=1in,height=1.25in,clip,keepaspectratio]{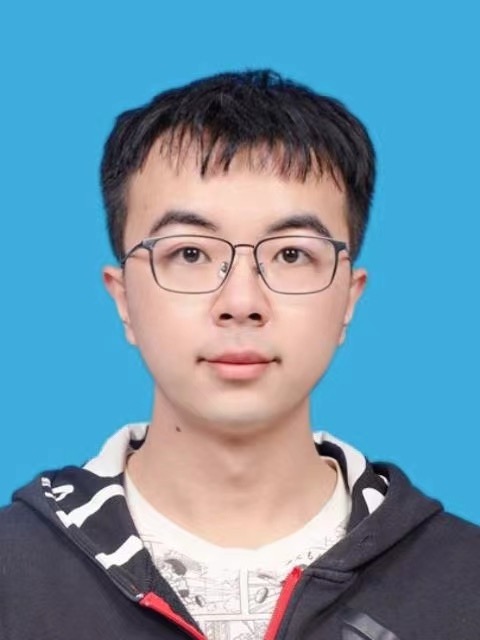}}]{Sixing Wu} received his Ph.D. degree in computer software and theory from Peking University, China, in 2022. He is currently a Lecture with the Software School, Yunnan University. His research interests include artificial intelligence, natural language processing, and open-domain dialogue response generation. He published 20+ academic papers in international conferences and journals such as ACL, EMNLP, IJCAI, COLING, TASLP, WWW, WSDM, and ICASSP. He also served as a PC member (reviewer) of several international conferences, such as ACL (and ARR), EMNLP, AAAI, etc.
\end{IEEEbiography}

\vspace{-10pt}

\begin{IEEEbiography}[{\includegraphics[width=1in,height=1.25in,clip,keepaspectratio]{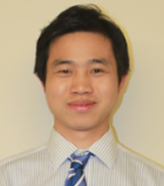}}]{Jun Zhao} (Member, IEEE) is currently an Assistant Professor in the College of Computing and Data Science at Nanyang Technological University (NTU), Singapore. He received a Ph.D. degree in Electrical and Computer Engineering from Carnegie Mellon University (CMU), Pittsburgh, PA, USA, in May 2015, and a bachelor’s degree in Information Engineering from Shanghai Jiao Tong University, China, in June 2010. One of his papers was a finalist for the best student paper award in IEEE International Symposium on Information Theory (ISIT) 2014. His research interests include A.I. and data science, security and privacy, control and learning in communications and networks.
\end{IEEEbiography}

\vspace{-10pt}

\begin{IEEEbiography}[{\includegraphics[width=1in,height=1.25in,clip,keepaspectratio]{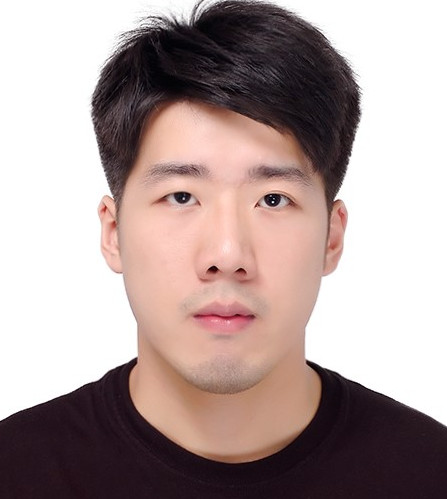}}]{Dongting Hu} is a Ph.D student in data science at the School of Mathematics and Statistics, University of Melbourne.  Prior to commencing PhD, he received his M.S. degree in Data Science from University of Melbourne in 2021. His research interests lie in 3D computer vision, including depth estimation and neural rendering.
\end{IEEEbiography}

\vspace{-10pt}

\begin{IEEEbiography}[{\includegraphics[width=1in,height=1.25in,clip,keepaspectratio]{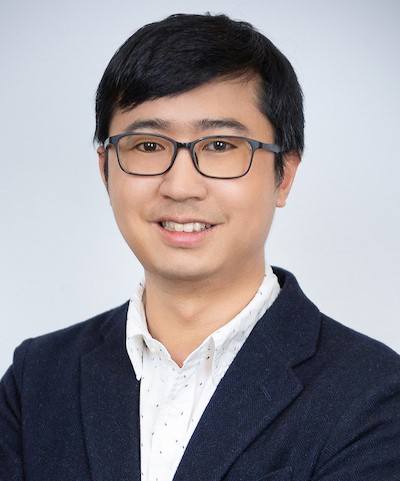}}]{Mingming Gong} is a Senior Lecturer in Data Science at the School of Mathematics and Statistics, University of Melbourne, and an affiliated Associate Professor of Machine Learning at the Mohamed bin Zayed University of Artificial Intelligence. His research interests include causal learning and inference, deep learning, and computer vision. He has authored and co-authored more than 50 research papers in top venues such as ICML, NeurIPS, ICLR, CVPR, TPAMI, IJCV, TIP, TNNLS and served as a meta-reviewer for NeurIPS, ICML, ICLR, IJCAI, AAAI, TMLR, etc. He has won the Australian Artificial Intelligence Emerging Researcher Award and the Australian Research Council Early Career Researcher Award.
\end{IEEEbiography}

\end{document}